\RequirePackage{tikz} 
\documentclass[sn-basic]{sn-jnl}
\usetikzlibrary{shapes.geometric}
\usetikzlibrary{matrix,positioning,shapes,shadows,shadows.blur,arrows,calc, fit,backgrounds}
\usetikzlibrary{intersections}

\usepackage{amsmath}
\usepackage{amssymb}
\usepackage{amsfonts}
\usepackage{booktabs}
\usepackage{mathtools}
\DeclarePairedDelimiter{\ceil}{\lceil}{\rceil}
\usepackage{subfig}
\usepackage{tabularx}
\usepackage{pifont}
\newcommand{\cmark}{\ding{51}}%
\newcommand{\xmark}{\ding{55}}%
\usepackage{enumerate}

\usepackage{xspace}

\newcommand{\mathsymbol}[2]{\newcommand{#1}{\ensuremath{\mathit{#2}}\xspace}}
\newcommand{\textmacro}[2]{\mathsymbol{#1}{\text{#2}}}

\mathsymbol{\learner}{a}
\mathsymbol{\learnerspace}{A}
\mathsymbol{\preexpfactor}{q}
\mathsymbol{\preexpbound}{\delta}

\mathsymbol{\sampleanchor}{n}
\mathsymbol{\timeanchor}{t}
\mathsymbol{\budget}{b}
\mathsymbol{\schedule}{S}

\mathsymbol{\dataspace}{\mathcal{D}}
\mathsymbol{\dataset}{d}
\mathsymbol{\trainset}{d_\mathit{tr}}
\mathsymbol{\instancespace}{\mathcal{X}}
\mathsymbol{\labelspace}{\mathcal{Y}}
\mathsymbol{\hypospace}{H}
\mathsymbol{\risk}{\mathcal{R}}
\mathsymbol{\lc}{f}
\mathsymbol{\lcab}{\lc(\learner, \budget)}
\mathsymbol{\lcest}{\hat \lc}
\mathsymbol{\lcestb}{\lcest_{\learner}(\budget)}
\mathsymbol{\lcestab}{\lcest(\learner, \budget)}
\mathsymbol{\prob}{\mathbb{P}}

\mathsymbol{\lcobservationset}{O}
\mathsymbol{\learningcurve}{\mathcal{C}}
\mathsymbol{\utilitycurve}{\mathcal{U}}

\mathsymbol{\curvemean}{\mu_{\cdot, \cdot}}
\mathsymbol{\curvemeana}{\mu_{\learner, \cdot}}
\mathsymbol{\curvemeanab}{\mu_{\learner, \budget}}
\mathsymbol{\curvenoisea}{\sigma_{\learner,\cdot}^2}
\mathsymbol{\curvenoiseab}{\sigma_{\learner,\budget}^2}

\mathsymbol{\satpoint}{\budget_{sat}}
\mathsymbol{\economicstoppingpoint}{\budget_{sat}^u}
\mathsymbol{\satperformance}{p_{sat}}
\mathsymbol{\limperformance}{p_{lim}}
\mathsymbol{\refpoint}{\budget_{\mathit{ref}}}
\mathsymbol{\refperformance}{p_{\mathit{ref}}}
\mathsymbol{\params}{\mathbf{\theta}}

\mathsymbol{\candidateranking}{\pi_{\learner \in \learnerspace}\sim \learningcurve(a, \refpoint)}

\textmacro{\samplecurve}{sample-wise curve}
\textmacro{\samplecurves}{sample-wise curves}
\textmacro{\itcurve}{iteration-wise curve}
\textmacro{\itcurves}{iteration-wise curves}

\newcommand*\rot{\rotatebox{90}}

\title[Learning Curves for Decision Making]{Learning Curves for Decision Making in Supervised Machine Learning: A Survey}

\author*[1]{\fnm{Felix} \sur{Mohr}}\email{felix.mohr@unisabana.edu.co}

\author[2]{\fnm{Jan N.} \sur{van Rijn}}\email{j.n.van.rijn@liacs.leidenuniv.nl}

\affil[1]{\orgdiv{\orgname{Universidad de La Sabana}, \city{Ch\'{i}a}, \state{Cundinamarca}, \country{Colombia}}}

\affil[2]{\orgdiv{Leiden Institute of Advanced Computer Science}, \orgname{Leiden University}, \orgaddress{ \city{Leiden}, \country{The Netherlands}}}

\definecolor{todocolor}{rgb}{1.0,0.2,0.2}
\definecolor{jvrcolor}{rgb}{0.8,0.2,0.8}
\definecolor{changedcolor}{rgb}{0.2,0.2,1.0}

\begin{document}

\abstract{
Learning curves are a concept from social sciences that has been adopted in the context of machine learning to assess the performance of a learning algorithm with respect to a certain resource, e.g., the number of training examples or the number of training iterations.
Learning curves have important applications in several machine learning contexts, most notably in data acquisition, early stopping of model training, and model selection.
For instance, learning curves can be used to model the performance of the combination of an algorithm and its hyperparameter configuration, providing insights into their potential suitability at an early stage and often expediting the algorithm selection process. 
Various learning curve models have been proposed to use learning curves for decision making.
Some of these models answer the binary decision question of whether a given algorithm at a certain budget will outperform a certain reference performance, whereas more complex models predict the entire learning curve of an algorithm. 
We contribute a framework that categorises learning curve approaches using three criteria: the decision-making situation they address, the intrinsic learning curve question they answer and the type of resources they use. 
We survey papers from the literature and classify them into this framework.}


\maketitle

\section{Introduction}\label{sec:introduction}
\begin{figure}[t]
    \centering
    \includegraphics[width=\textwidth,trim=1.5cm 0cm 1cm 0,clip]{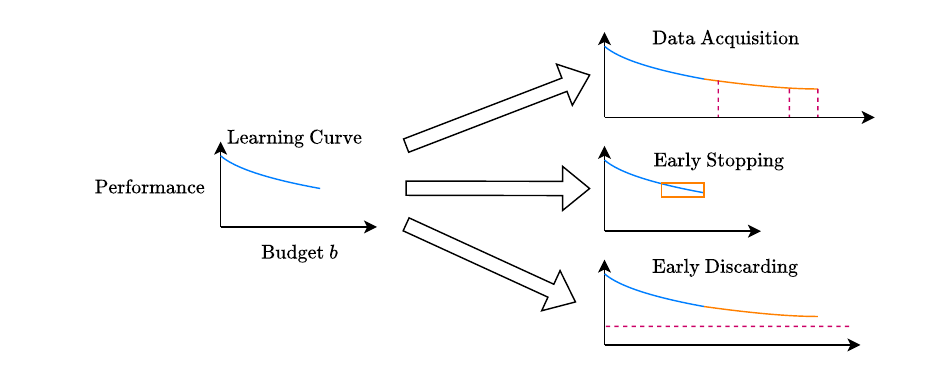}
    \caption{The three types of decision-making situations in which learning curves are typically used. The $x$-axis of each figure represents the budget in the applicable unit, and the $y$-axis represents the performance.}  \label{fig:decision_situations}
\end{figure}

Learning curves describe a system's performance on a task as a function of some resource to solve that task.
There can be a pre-defined budget of that resource, limiting the amount of resources that can be spent. 
In other cases, the goal can be to obtain reasonable results while minimising the spent budget of that resource. 
Typical types of budgets are the number of \emph{examples} the learner has observed before performing the task or the number of \emph{iterations} or \emph{time} the learner spends in an environment.
The performance measure expresses the quality of the obtained model, e.g., error rate or F1 measure.
Learning curves are an important source of information for making decisions on the following matters in machine learning:

\begin{itemize}
    \item \emph{Data Acquisition} determines how many data points should reasonably be acquired to obtain a desired performance. The top right plot in Fig.~\ref{fig:decision_situations} visualises a scenario where we have already observed performance up to a certain amount of data (the blue learning curve). We can extrapolate this and make a prediction of what the performance would be if more data was available, i.e., the value of the orange extrapolation at different vertical pink lines in the figure \citep[see, e.g.,][]{weiss08maximizingclassifierutility,last2009improving}. 
    
    \item \emph{Early Stopping} of training a model.
    If we are committed to some specific learner (a learning algorithm \emph{and} its hyperparameters), we might want to minimise the training time~\citep{provost1999efficientprogressivesampling,john1996staticvsdynamic} or avoid over-fitting~\citep{bishop1995regularization,goodfellow2016deep}.
    The middle right plot in Fig.~\ref{fig:decision_situations} visualises a scenario where we have already observed performance up to a certain amount of budget, and based on the progression of the learning curves on recent iterations, a decision can be made whether to continue the learning or terminate it.
    \item \emph{Early Discarding} in model selection.
    If we want to \emph{select} from various models, we want to stop the evaluation of a candidate when we are reasonably certain that it is not \emph{competitive} to the best-known solution~\citep{domhan2015speedingup,swersky2014freezethawbo,mohr2023fast}. The bottom right plot of Fig.~\ref{fig:decision_situations} visualises a scenario where we have already observed performance up to a certain amount of budget (the blue learning curve) and already an incumbent performance obtained by an earlier configuration (horizontal dashed pink line). By using learning curve extrapolation techniques, we can determine whether the current configuration can surpass the incumbent configuration; if not, discarding the current training process early (as in the case shown) would be justified.
\end{itemize}
Many techniques with varying complexity and required resources have been proposed to address either of these problems.
The complexity ranges from approaches that simply recognise whether an already observed part of a learning curve has converged \citep{bishop1995regularization,provost1999efficientprogressivesampling} to the creation of parametric learning curve models, which capture a belief model for the behaviour of various learners at any possible budget~\citep{klein2017lcpredictionwithbnn}.
While simple approaches may only rely on the observations made so far for a learner on the dataset of interest, more complex approaches may rely on additional resources such as learning curves or features of other datasets \citep[see, e.g.,][]{leite2010activetesting} or learners \citep[see, e.g.,][]{chandrashekaran2017speedinguphpo} or both \citep[see, e.g.,][]{ruhkopf2022masif}.

Our contribution is a unified framework of the usage of learning curves for decision making in machine learning and an extensive review of the literature of approaches that fall within this framework.
This framework categorises the existing literature along the following three axes: 
\begin{enumerate}
\item The type of decision-making situation, i.e., whether it is used to make decisions about data acquisition, early stopping, or early discarding. See Sec.~\ref{sec:taxonomy:situations} for more details.
\item The type of technical question that can be answered with an approach, e.g., some approaches can only answer the binary question whether a model has converged,
whereas other approaches are able to answer questions about the behaviour of any part of the learning curve. See Sec.~\ref{sec:taxonomy:technicalquestions} for more details.
\item The data resources that are used to model the learning curve. For example, in some cases, data from different algorithms on the same dataset is being used, whereas in other cases, data from the same algorithm on other datasets. See Sec.~\ref{sec:taxonomy:resources} for more details.
\end{enumerate}
We perform an extensive literature survey in which we categorise published learning curve extrapolation models along the various axes of this framework.
This literature survey is subject of Sec.~\ref{sec:literature}, which is then summarised in Table~\ref{tab:overview}.
We focus specifically on supervised machine learning, in which learning curves describe the predictive performance of a model produced by a learning algorithm either as a function of the number of training instances or of the time or iterations spent for learning on a given dataset.
We explicitly exclude learning curves that describe the performance of a learning agent in an environment over time, i.e., the learning curve of an agent in a reinforcement learning setup~\citep{waltz1965heuristic}.
Similarly, we briefly contrast learning curves to other performance curves, such as active learning curves, feature curves, and capacity curves, and explain why we consider these out of scope for the literature review.
Still, we aim to survey exhaustively the literature that introduces approaches that use learning curves in supervised learning.

\paragraph{Contributions:} Our contributions are the following. 
\begin{itemize}
\item We present a unified framework of the usage of learning curves for decision making in machine learning and an extensive review of the literature on approaches that fall within this framework. This framework contains three axes, i.e., (i)~the type of decision-making situation, see also Fig.~\ref{fig:decision_situations} (ii)~the type of question that can be answered, see also Fig.~\ref{fig:questions:learningcurve} and (iii)~the data resources that are used to model the learning curve, see also Fig.~\ref{fig:resources}. While the first axis of this framework is also used in other literature \citep[see, e.g.,][]{viering2023theshape}, to the best of our knowledge, the other two axes have not yet been explicitly identified. 

\item We conducted a literature survey in which we categorise methods presented in the literature along the various axes of this framework. Sec.~\ref{sec:literature} lists all these methods (where each subsection represents a type of question being answered), and Table~\ref{tab:overview} overviews all methods along the three axes of our framework. 
\item Based on the framework, we identify unexplored routes for further research. Most notably, we note that there is a mismatch between the research questions being answered and the learning curve modelling method being used; in many cases, a high-level modelling technique is used to answer a low-level question. We speculate that matching the level of the question being answered with the appropriate level of the modelling technique can further improve the obtained results. 
\end{itemize}
\paragraph{Relation to other literature reviews on learning curves:} Another prominent literature review that centres around learning curves is the highly complementary work by \citet{viering2023theshape}, which has been developed in parallel. While both works have identified the three types of decision-making situations that are referred to in the literature, \citet{viering2023theshape} survey more theoretical work that analyses the shape of learning curves, whereas this work surveys work that is more oriented towards methods that extrapolate learning curves, thereby supporting the data scientists in various decision-making situations.

\paragraph{Structure:} This paper is structured into three main parts. 
Sec.~\ref{sec:background} presents relevant background knowledge on learning curves, including formal definitions and the terminology relevant for the remainder.
Sec.~\ref{sec:curvemodels} presents relevant important concepts that relate to how learning curves are generally modelled.
Sec.~\ref{sec:taxonomy} contains our main contribution by introducing our framework for categorising methods that utilise learning curves for decision making in supervised learning. 
Following this framework, Sec.~\ref{sec:literature} exhaustively reviews approaches that explicitly or implicitly answer questions related to learning curves to make or recommend decisions in the context of supervised machine learning.
Sec.~\ref{sec:conclusions} concludes our findings.
Finally, Appendix~\ref{app:notation} presents a table that overviews the most critical notation used throughout this paper.

\section{Background on Learning Curves}
\label{sec:background}
This section gives a conceptual background on learning curves.
It first provides an idealised formal definition in Sec.~\ref{sec:background:definition} followed by a definition of \emph{empirical} learning curves in Sec.~\ref{sec:background:empiricalcurves} that can be computed in practice.
The concept of \emph{utility} curves is introduced in Sec.~\ref{sec:background:utilitycurves}.
Sec.~\ref{sec:background:shapes} introduces important terminology such as \emph{anchor points}, \emph{limit performance}, and the \emph{saturation point}.
Finally, Sec.~\ref{sec:background:othertypes} contrasts the learning curves covered in this survey with other types of performance curves used in machine learning.

\subsection{Sample-Wise and Iteration-Wise Learning Curves}
\label{sec:background:definition}
We consider learning curves in the context of supervised machine learning.
Formally, in the supervised learning context, we assume some \emph{instance space} \instancespace and a \emph{label space} \labelspace.
A \emph{dataset} $\dataset \subset \{(x,y)~\vert~x\in \instancespace, y \in \labelspace\}$ is a \emph{finite} relation between the instance space and the label space.
We denote as \dataspace the set of all possible datasets.
A \emph{learning algorithm} is a function $\learner: \dataspace \times \Omega \rightarrow \hypospace$, where $\hypospace = \{h~\vert~h: \instancespace \rightarrow \labelspace\}$ is the space of hypotheses and $\Omega$ is a source of randomness.

Note that learning curves can also be considered in other machine learning setups.
In fact, learning curves appeared first in reinforcement learning~\citep{waltz1965heuristic} and have also been used for unsupervised learning~\citep{meek2002thelcsamplingmethod}.
However, to give this survey focus, we consider learning curves for supervised learning.

The performance of a hypothesis is typically expressed as \emph{risk}, which is also often called \emph{out-of-sample error}:
\begin{equation}\label{eq:risk}
\risk_{out}(h) = \int\limits_{\instancespace,\labelspace}{loss(y, h(x))} \, d \prob_{\instancespace\times\labelspace}.
\end{equation}
Here, $\mathit{loss}(y,h(x)) \in \mathbb{R}$ is the penalty for predicting $h(x)$ for instance $x \in \instancespace$ when the true label is $y \in \labelspace$, and $\prob_{\instancespace\times\labelspace}$ is a joint probability measure on $\instancespace \times \labelspace$ from which the available dataset \dataset has been generated.
As such, the out-of-sample error represents the weighted summed error that hypothesis $h$ makes on all possible instance-label pairs, weighted by their probabilities.

The performance of a \emph{learning algorithm} is simply the performance of the hypothesis it produces.
In contrast to the performance of a hypothesis, the performance of a learner depends on its input, i.e., on the data provided for learning.
The average performance of learner \learner for a number \sampleanchor of training examples can then be expressed as
\begin{equation}
    \label{eq:learningcurve:def1}
    \learningcurve(\learner, \sampleanchor) = \int\limits_{\mathclap{\omega \in \Omega, \trainset \in \dataspace, \vert\trainset\vert = \sampleanchor}}
    \risk_{out}(\learner(\trainset, \omega))d\prob_{\instancespace\times\labelspace} d\prob_\Omega,
\end{equation}
where $\trainset \in \dataspace$ is the dataset of size \sampleanchor used to induce a model using learner \learner.
It is generally assumed that \dataset is a collection of i.i.d. samples from $\prob_{\instancespace\times\labelspace}$.

\begin{figure}[t]
    \centering
    \includegraphics[width=.4\textwidth,trim=0.9cm 0 20cm 0,clip]{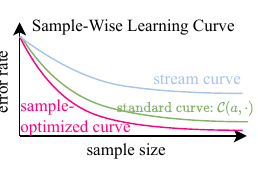}
    \hspace{1cm}
    \includegraphics[width=.4\textwidth,trim=6cm 0 15cm 0,clip]{fig/performancecurves.pdf}
    \caption{Left: (Standard) \samplecurve (green) for a single learner on a particular data source together with learning curves under sample optimisation (pink) and learning curves on streams (blue).
    Right: \itcurves of a single learner on a particular data source for two different dataset sizes $n_1 < n_2$.\label{fig:performancecurves_vanilla}}
\end{figure}

When we consider Eq.~(\ref{eq:learningcurve:def1}) as a function of the number of training samples for a fixed learner \learner, we obtain the \emph{\samplecurve} of learner \learner.
That is, the \samplecurve is the function $\learningcurve(\learner, \cdot): \mathbb{N} \rightarrow \mathbb{R}$; so it is a \emph{sequence} of performances, one for each training size.
Fig.~\ref{fig:performancecurves_vanilla} (left) visualises this by means of the green line and compares this to two other types of learning curves with the error rate as the loss (see Sec.~\ref{sec:background:othertypes}).

Alternatively, many learning algorithms implement an iterative internal optimisation process, which allows describing the learning progress over time or a number of iterations.
For example, neural network training produces a new hypothesis after every batch or epoch, ensemble learners like bagging or boosting produce a new hypothesis after every added ensemble member, and support vector machine optimizers yield updated attribute or instance coeffients in iteration.
In the formal framework, a learner can be seen more generally as a function $\learner: \dataspace \times \Omega \rightarrow \hypospace^+$ that maps a dataset to a \emph{sequence} of hypotheses, one for each of its iterations.
The above error function for learners can then be written as
\begin{equation}
    \label{eq:learningcurve:def2}
    \learningcurve(\learner, \sampleanchor, \timeanchor) = \int\limits_{\mathclap{\omega\in \Omega, \trainset \in \dataspace, \vert\trainset\vert = \sampleanchor}}\risk_{out}(\learner(\trainset, \omega)_\timeanchor)d\prob_{\mathcal{X}\times \mathcal{Y}} d\prob_\Omega,
\end{equation}
Here, \timeanchor expresses some budget, for example, time or a number of iterations over the dataset, often expressed in epochs.

Based on this notion, the \emph{\itcurve} of a learner \learner is defined for a \emph{fixed} dataset size \sampleanchor (often between 70\% and 90\% of the available data) and is then the function $\learningcurve(\learner, \sampleanchor, \cdot): \mathbb{N} \rightarrow \mathbb{R}$.
Fig.~\ref{fig:performancecurves_vanilla} (right) visualises an example of two \itcurves. It can be seen that these \itcurves are also influenced by the number of samples. The \samplecurve in this example (visualised by the dashed line) is assumed to utilise the maximal number of iterations.
Examples of such learning curves occur above all in the analysis of deep learning models~\citep{domhan2015speedingup,goodfellow2016deep}.

The two types of learning curves seem to be related and indeed look similar when visualised, but they have different semantics.
The crucial difference is that \itcurves are usually visualized for a fixed and finite number of training samples (expressed by \sampleanchor), no matter how large \timeanchor becomes.
In fact, since iterative learners typically automatically stop the learning process as soon as no progress is observed, it often holds that $\learningcurve(\learner, \sampleanchor) = \lim_{\timeanchor \rightarrow \infty} \learningcurve(\learner, \sampleanchor, \timeanchor)$.
But this is not necessarily the case, specifically if an algorithm stops early (in the \itcurve), e.g., to avoid over-fitting (as the neural network in Fig.~\ref{fig:lcs_waveform}).
Note that as \timeanchor grows, each training instance is considered an unlimited number of times; hence, \itcurves show how much the learner can make out of a constant number of training instances.
Instead, \samplecurves show the performance of the learner as the number of \emph{examples} grows.
The latter typically means, for an infinite input space, that the \emph{information basis} available to the learner is growing strictly bigger, while the information is constant in the case of \itcurves.

Note that while the learning success can be expressed in a metric that ought to be maximised or minimised, in this paper we assume for simplicity that they are to be minimised.
This is why the performance is expressed through a \emph{loss} such as the error rate.
In this case, learning curves are (usually) \emph{decreasing}.
However, more generally, one can also be interested in \emph{increasing} learning curves, e.g., when considering accuracy or the F1 measure.
Since learning curves can be simply mirrored at the x-axis, every approach discussed in this paper is applicable to both increasing or decreasing learning curves.
For simplicity, this survey assumes that lower performance values are better (error rate, log-loss/cross-entropy, Brier score, mean-square-error, etc.).

\subsection{Empirical Learning Curves}
\label{sec:background:empiricalcurves}
The above definitions of learning curves are purely theoretical.
This is because we cannot evaluate equations (\ref{eq:risk}-\ref{eq:learningcurve:def2}) in practice.
First, the out-of-sample error $\risk_{out}$, i.e., Eq.~(\ref{eq:risk}) cannot be computed in practice since the measure $\prob_{\instancespace\times\labelspace}$ is unknown.
Relying on this error, the learning curve values cannot be computed either.
The necessity to average over the oftentimes uncountable set of all possible train sets can add additional problems.

\begin{figure}
    \centering
    \includegraphics[width=\textwidth]{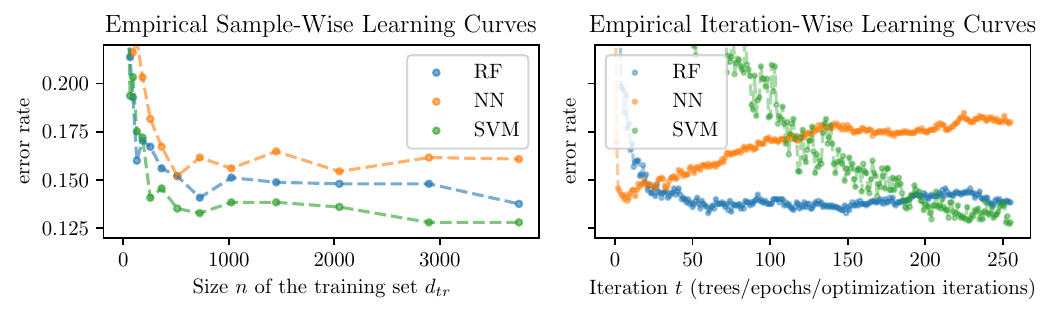}
    \caption{Empirical learning curves for the waveform dataset. Left: \samplecurves at different training set sizes up to 80\% of the data.
    The remaining 20\% are used to compute the error.
    Right: \itcurves with one entry for each forest size (RF), epoch (NN), or optimization iteration (SVM) when a fixed set of 80\% of the available data is used in each iteration for training and the rest to compute the error.}
    \label{fig:lcs_waveform}
\end{figure}

To compute learning curves in practice, we rely on empirical \emph{estimates} of the above quantities.
We estimate the out-of-sample error by the \emph{internal error}:
\begin{equation}\label{eq:insampleerror}
\risk_{in}(h) = \frac{1}{\vert\dataset\vert}\sum_{(x,y)\in\dataset} loss(y, h(x)),
\end{equation}
where \dataset is the dataset used for assessing the performance of hypothesis $h$.
Note that the dataset \dataset may or may not contain instances used to create the hypothesis $h$.
In most practical applications, dataset \dataset consists of instances that have not been seen during the creation of hypothesis $h$ (i.e., the test set). However, in theory, the internal error can also be estimated based on the train set \trainset or a combination of instances from the train set and test set.

We consider an \emph{empirical learning curve} any set of estimates of a true learning curve for different sizes or iterations.
We can use various estimation procedures to estimate the performance using a given training size, such as using a regular holdout set or cross-validation. The latter leads to various estimates, and
averaging over these estimates yields an estimate of the \samplecurve in Eq.~(\ref{eq:learningcurve:def1}) at size \sampleanchor.
To obtain an empirical estimate of the iterative learning curve in Eq.~(\ref{eq:learningcurve:def2}), we do the same except that we stop the learning algorithm after \timeanchor iterations.

Fig.~\ref{fig:lcs_waveform} shows empirical learning curves for a Random Forest (RF), a Neural Network (NN) with 100 neurons in a hidden layer, an a support vector machine (SVM) with RBF kernel on a concrete and widely used benchmark dataset (the waveform dataset).
The empirical curves are the scatter points; the lines here are only a visual aid.
The error rate is here obtained from a single validation fold, i.e., without averaging, which explains the rather unsmooth behavior.
The \itcurves were created using 80\% of the data for training.

Since empirical learning curves are the only way to gain insights about true learning curves, quite some studies have been published with the sole goal of sharing empirical learning curves with the community and thereby improving the understanding of how they behave.
\citet{perlich2003treeinduction} contrast the learning curves on decision trees and logistic regression on different datasets.
Notably, the authors also \emph{compare} learning curves, e.g., they report whether one curve is below the other (dominates it) on all considered training set sizes or whether the two learning curves cross.
Several other studies report learning curves for specific learners.
\citet{ng2001ondiscriminative} compare logistic regression and naive Bayes.
\citet{morch1997nonlinear} conduct a study similar to the one by \citet{perlich2003treeinduction} to compare linear vs. non-linear classifiers on a smaller scale.
Recently, a number of learning curve databases have been published, i.e., learning curves of different network architectures on typical image classification datasets \citep{bornschein2020smalldatabigdecisions,nas201,nas301}, learning curves of different machine learning algorithms on tabular data \citep{lcdb}, and for mixtures of these tasks  \citep{yahpo,hpobench}.

Empirical studies of this type do not answer generalising questions about learning curves but rather report experimental results.
This is different from contributions in which certain model assumptions are made and data is compared to those models, e.g., with the aim to compute the goodness of fit of that model.
In Sec.~\ref{sec:curvemodels}, we briefly discuss some of such models.

\subsection{Utility Curves}
\label{sec:background:utilitycurves}

The concept of learning curves can be further generalised to a \emph{utility curve}~\citep{last2007predicting,weiss08maximizingclassifierutility,last2009improving}.
The utility usually involves a trade-off between the performance and the computational cost of training a model. The specific details can be different per task.
The utility is connected to the learning curve in so far as the utility is also a function of the budget and is directly influenced by the predictive performance.
Therefore, one could argue that the utility curve \utilitycurve is obtained by passing the learning curve to the utility function alongside other parameters that influence the utility, most notably the cost of acquiring new instances and the cost to train a model on the respective dataset size.
The learning curves associated with utility costs are visualised in Fig.~\ref{fig:performancecurves_utility} (orange) and compared to a normal learning curve. 
Assuming there is a linear cost associated with further training the classifier, we can see that the utility curve (which makes a trade-off between performance and cost) peaks at a given point, and deteriorates afterwards.

        \subsection{Terminology of Learning Curves}
\label{sec:background:shapes}
\begin{figure}[t]
    \centering
    \includegraphics[width=.95\columnwidth]{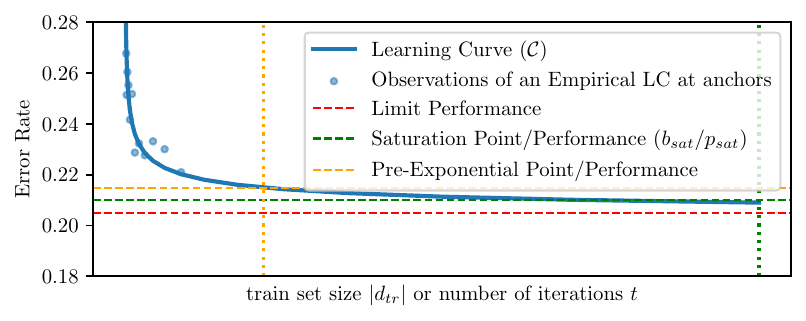}
    \caption{Concepts related to a learning curve. Red: Limit performance. Green: Saturation point \satpoint (vertical line) and saturation performance \satperformance (horizontal line). Orange: Pre-exponential point (vertical line) and pre-exponential performance (horizontal line).
    The curve plateaus at a performance of 0.2; the plateau is not visualized here to emphasise the difference between the pre-exponential point and saturation point.
    }
    \label{fig:lcconcepts}
\end{figure}

While this survey is not primarily about the shapes of learning curves, the shapes of learning curves play an important role when using them to make decisions.
Hence, we consider it necessary to convey some of the most important insights about the basics of the shapes of learning curves.
However, we refer to a recent exhaustive survey on the shapes of learning curves~\citep{viering2023theshape} for details on this topic. Fig.~\ref{fig:lcconcepts} visualises some important concepts.

\begin{figure}[t]
    \centering
    \includegraphics[width=.4\textwidth,trim=10cm 0 10cm 0,clip]{fig/performancecurves.pdf}
    \caption{A utility curve with the corresponding learning curve. \label{fig:performancecurves_utility}}
\end{figure}

\paragraph{Anchor Points}
In this survey, we adopt the term \emph{anchor} to refer to a point for which the empirical learning curve carries a performance estimate.
There is no established name for such points in literature.
They are called \emph{sample sizes} in~\citep{john1996staticvsdynamic,provost1999efficientprogressivesampling,leite2007aniterativeprocess}, and those authors refer to a \emph{collection} of such samples sizes as a \emph{schedule}~\citep{john1996staticvsdynamic,provost1999efficientprogressivesampling,figueroa2012predictingsamplesize,meek2002thelcsamplingmethod}.
However, for \itcurves, the term `sample' is misleading because the curve plots performance against the number of times all instances (out of a fixed set) are presented to the learner, which is not the same as sample size.
Besides, the term sample size is quite overloaded in the context of machine learning, because this field deals with various types of samples in different contexts, e.g., train and validation samples, etc.
Another terminology observed sometimes is the one of \emph{sample landmarks}~\citep{furnkranz2001evaluation,leite2005predictingrelativeperformance}.
However, this term is also slightly confusing, since landmarks are generally understood as the performances of cheap-to-evaluate learners, which was also the motivation for this terminology by \citet{furnkranz2001evaluation}.
A less used terminology is the term \emph{anchor}~\citep{kolachina2012predictioninmachinetranslation,mohr2023fast,mohr2021towards,kielhofer2024learning}, which is not ambiguous in the machine learning context and captures the idea that analysis is based on some selected points.
It serves well to immediately create an association with a particular size of a sampled training data set or a number of visited instances that is used in the context of building an empirical learning curve.

Throughout this paper, we formally use the symbol \budget to refer to an anchor.
Thereby, we abstract away from sample sizes \sampleanchor or iteration \timeanchor.
In other words, the symbol \budget is used to indicate points on both \samplecurves or \itcurves, and it should be clear from the context which one is meant (if the difference is relevant).

For example, in Fig.~\ref{fig:lcs_waveform} above, we have the following anchors.
For the \samplecurve, a geometric schedule with $n \in \{\ceil{2^{i/2}}~\vert~ i \in \mathbb{N}\}$ is chosen.
In this concrete case, the anchors are \{1, 2, 3, 4, 6, 8, 12, 16, 23, 32, 46, 64, 91, 128, 182, 256, 363, 512, 725, 1024, 1449, 2048, 2897, 3750\}.
In the \itcurve, every iteration is used as an anchor, so $t \in \{1,..,200\}$.

\paragraph{Limit Performance}
It is generally assumed that learning curves \emph{converge} to some value.
In the case of iterative learning curves, there are sometimes oscillations in the curve, but even in such cases, the curve usually converges to some (in those cases, typically a bad) value eventually.
We are not aware of a particular term that is used to describe the score to which a learner converges.
\citet{cortes1994limits} describes the \emph{limiting performance} or the \emph{asymptotic performance} of the \emph{data}; i.e., it is not the property of a particular learner but the best achievable performance among all learners under consideration (even though only tested with two model types in the paper).
In this paper, we adopt the term \emph{limit performance} of the learner (on a fixed number of training instances in the case of \itcurves), and we denote this performance as \limperformance.

\paragraph{Saturation Point}
Intuitively, the \emph{saturation point} is the anchor after which the performance convergences.
That is, the anchor after which all values are in a distance of less than some pre-defined and typically very small $\varepsilon$.
In early works, this point has been called \emph{stopping point}~\citep{provost1999efficientprogressivesampling,figueroa2012predictingsamplesize,meek2002thelcsamplingmethod,leite2003improvingprogressivesampling,leite2004improvingprogressivesampling}.
\citet{provost1999efficientprogressivesampling} characterise this point as follow: ``Models
built with smaller training sets [than $s_{min}$] have lower accuracy than models built
with from training sets of size $s_{min}$, and models built with larger training sets have no higher accuracy.''
In the context of those works, namely progressive sampling, the term stopping point makes sense because they progressively sample until they reach the convergence region, and then stop the sampling procedure.
In the absence of such a mechanism, the term appears a bit odd.
The notion of saturation in the context of learning curves was proposed by \citet{tomanek2010resourceawareannotation} and seems appropriate since the limit performance has not been reached, but it has \emph{almost} been reached.
The curve is saturated up to a mistake of $\varepsilon$.

We will denote the saturation point itself as \satpoint, emphasising that the saturation point is an anchor.
The \emph{saturation performance} is simply the value of the learning curve at that point.
Both are defined in the context of a fixed learner \learner.
Hence, we can write the saturation performance as $\satperformance := \learningcurve(\learner, \satpoint)$ in the context of a \samplecurve or $\satperformance := \learningcurve(\learner, \sampleanchor,\satpoint)$ in the context of an \itcurve.

\paragraph{Pre-Exponential Point}
A related concept is the pre-exponential point and, correspondingly, the pre-exponential performance.
The saturation point may be expensive to reach in the sense that a lot of training data is necessary to obtain the saturation performance.
We call the smallest anchor point for which an increase by a factor of \preexpfactor leads to a performance improvement of less than some \preexpbound, i.e., $\learningcurve(\learner, \sampleanchor) - \learningcurve(\learner, \preexpfactor \cdot \sampleanchor) < \preexpbound$.
Reasonable candidates for \preexpfactor can be 2 or 10, while candidates for \preexpbound can be $0.01$ or $0.001$ if the metric is the error rate.
Its semantic is that from the pre-exponential point on, one needs more than $\preexpfactor^k$ (i.e., an exponentially increasing number of) training samples or iterations to improve by a low margin of $k \preexpbound$.
Correspondingly, the pre-exponential performance is the performance that can be obtained by a comparably small anchor point.

\paragraph{Utility-Based Stopping Point}
The \emph{utility-based stopping point} is the point at which the acquisition of further data points has a negative impact on the \emph{utility} of the data analysing entity.
Therefore, this concept is associated with utility curves. 
The utility-based stopping point is not related to the saturation point but, if at all, rather to the pre-exponential point; we denote it as \economicstoppingpoint.
In Fig.~\ref{fig:performancecurves_utility}, this is the peak of the yellow curve.

\paragraph{Plateau}
The right-sided open interval bounded by the saturation point from the left is consistently called the \emph{plateau} of the curve.
However, a curve can also have \emph{intermediate} plateaus, i.e., intervals of (almost) constant performance without being the final plateau.

\paragraph{Well-behaved Learning Curves}
To our knowledge, the notion of a \emph{well-behaved} learning curve is first used by \citet{provost1999efficientprogressivesampling}.
A learning curve is said to be well behaved if its slope is monotonically non-decreasing (for error-based learning curves).
An even stricter criterion demanding \emph{convexity} of the curve, which implies monotonicity, has been introduced recently by \citet{mohr2023fast,mohr2021towards}.
The property of being well behaved is one of the \emph{true} learning curve.
The (linear interpolation of an) empirical learning curve can often violate this condition, specifically when the number of validations conducted at the anchors is small or when the learning curve has reached a plateau.

While it is known that not all learning curves are well behaved~\citep{loog2012dipping,loog2019minimizers}, empirical studies suggest that such curves are rather an exception and that most \emph{\samplecurves} are well behaved.
Learning curves that are not well behaved are found above all in the context of deep learning, where a \emph{double descent} or \emph{peaking} phenomenon can be observed (at times) for both \samplecurves and \itcurves~\citep{nakkiran2020deepdoubledecent}; the effect was also observed decades ago for other learners~\citep{vallet1989linear}.
However, extensive empirical studies have shown that most \samplecurves are even \emph{convex}~\citep{mohr2023fast} and hence well behaved.
Some recent works suggest that potential ill behaviour can be mitigated by appropriate configuration or wrapping of learners~\citep{nakkiran2020optimalregularization,viering2020makinglearnersmonotone,mhammedi2020risk}.

\subsection{Relation to Other Types of Performance Curves}
\label{sec:background:othertypes}
We briefly discuss the relation of learning curves to other types of curves and problem settings.
These curves are fundamentally different from learning curves, and therefore, a more detailed coverage is beyond our scope.

\subsubsection{Learning Curves in Active Learning}
Active learning is a setting where the data scientist can acquire the label of arbitrary instances~\citep{settles2009active} and hence can actively increase the training set.
On a concrete data source with a concrete initial dataset, a specific active learning strategy creates a \emph{deterministically} extended dataset for any arbitrary anchor.
While this allows drawing a curve that plots performance against the number of training instances, this curve is not the one described in Eq.~(\ref{eq:learningcurve:def1}), because the datasets are not sampled i.i.d. from $\mathcal{P}_{\instancespace \times \labelspace}$ but dictated by the active learning strategy; they are \emph{sample-optimised}.
Fig.~\ref{fig:performancecurves_vanilla} displays how this theoretically relates to the normal \samplecurve.
We do not consider this type of learning curve in this survey.

\subsubsection{Learning Curves under Optimal Class Distribution}
\label{sec:literature:classdistribution}
Similarly, we obtain such a biased learning curve if we do not preserve the \emph{class} distribution.
\citet{weiss2003learningwhentrainingdatarecostly} have shown that it can be advantageous to over-sample instances of a minority class if they occur substantially more seldom than instances of a majority class.
One can then ask for the best class distribution for a certain anchor.
Similar to the active learning case, this creates a new distribution of datasets that does not coincide with $\mathbb{P}_{\instancespace \times \labelspace}$ anymore.
If we optimise over the class distribution at each anchor, we obtain a curve with the same axis labels as a \samplecurve but a different semantics (and most likely different values).
For the case of two classes, this type of learning curve is obtained by taking the budget-wise maximum of a performance \emph{surface} as proposed by \citet{forman2004learningfromlittle}.

\subsubsection{Learning Curves on Data Streams}
Another type of learning curve that violates the implicit assumptions made in Eq.~(\ref{eq:learningcurve:def1}) is obtained when learning from \emph{data streams}, a scenario in which training instances are coming in sequentially and need to be processed under strict time and memory constraints~\citep{bifet2018machine}.
Incremental learners like Hoeffding trees~\citep{domingos2000mining} and models induced by stochastic gradient descent are natural solutions to this problem domain.
When applied to a data stream, these algorithms explicitly \emph{forget} an instance once it has been processed.
This is not only to free memory but also to address the problem of \emph{concept drift}, i.e., the fact that $\prob_{\instancespace \times \labelspace}$ changes over time.
In such a case, we do not have that $\learningcurve(\learner, \sampleanchor) = \lim_{\timeanchor\rightarrow \infty} \learningcurve(\learner, \sampleanchor, \timeanchor)$ but rather that $\learningcurve(\learner, \sampleanchor) = \learningcurve(\learner, \sampleanchor, \sampleanchor)$, where the learner is updated at every incoming training instance, and each of the \sampleanchor instances was considered exactly once; for example, similar to training a neural network for one epoch with batch size 1.
While this produces a kind of \samplecurve, the result is clearly different from the learning curve received in the normal batch setting.
Fig.~\ref{fig:performancecurves_vanilla} displays how these type of curves theoretically relate to the normal \samplecurves.

Even though syntactically equivalent to a \samplecurve, data stream learning curves are substantially different and need different treatment.
Due to the (potential) concept drift over the time dimension, the i.i.d. assumption is not guaranteed~\citep{da2016using}.
From a theoretical viewpoint, extrapolating the learning curve over time to meaningfully predict future behaviour becomes impossible without the i.i.d. assumption.
Due to its special nature, the data stream setting is beyond our scope.

\subsubsection{Feature Curves}
\label{sec:background:featurecurves}
Learning curves always consider a fixed number of features.
Instead, one can fix the number of training instances and consider the performance as a function of the number of features.
This yields so-called \emph{feature curves}~\citep{hughes1968featurecurves1,viering2023theshape}.
\begin{figure}[t]
    \centering
    \includegraphics[width=.4\textwidth,trim=15cm 0 5.6cm 0,clip]{fig/performancecurves.pdf}
    \caption{Feature curve example for a single learner, once for a fixed and finite dataset size and once for an infinite dataset size. \label{fig:performancecurves_feature}}
\end{figure}

Defining meaningful feature curves is conceptually more difficult than learning curves because of the importance that single features can have.
For simplicity, consider only \samplecurves for this comparison.
In such learning curves, every point \sampleanchor is associated with the expected performance when using \sampleanchor training instances.
These \sampleanchor training instances are assumed to be drawn independently and identically distributed from the underlying distribution.
According to the aforementioned definitions of a learning curve (per Eq.~\ref{eq:learningcurve:def1} and Eq.~\ref{eq:learningcurve:def2}), there is no notion of a more informative instance (even though this notion clearly exists in the field of \emph{active learning}).
In particular, the \emph{order} in which instances are drawn is irrelevant.
However, in the context of features, some features are often more informative than others (and again, other features that have not been measured may be even more informative).
Therefore, in order to get an adequate overview, feature curves require some kind of averaging over all possible sets of features of a fixed size that can be formed from a base set of (available) features, as done by \citet{hughes1968featurecurves1}.
The fact that some features might be more important than other features makes it hard to model and extrapolate feature curves, as there is no reasonable set of assumptions to build these models on.
Fig.~\ref{fig:performancecurves_feature} displays two theoretical examples of feature curves. When having a finite number of samples, the performance of these curves will, in the limit (when more features are added), deteriorate due to the curse of dimensionality. Of course, when having infinite samples, the performance of such feature curves will go to perfect performance.

Note that feature curves and learning curves can be integrated.
For example, \citet{strang2018dontruleout} look at the combination of the number of instances and the number of features.
Since the effect of the number of features and the number of instances on the overall performance is clearly not independent, considering both together is sensible.
At the same time, due to the ambiguous semantics of feature curves already discussed above, using such combined curves is not necessarily straightforward for decision making, and we are not aware that the combined curve has been used for decision making so far.

\subsubsection{Capacity Curves}
\citet{cortes1994limits} introduce a curve that plots the performance of a configurable learner as a function of the complexity of its instantiation.
For example, the learner could be a neural network, and the complexity would then be the number of hidden layers.
In doing this, a fixed dataset size is assumed.
In that paper, this type of curve has no specific name, but we dub it the \emph{capacity curve} because they plot the performance as a function of capacity.
Fig.~\ref{fig:performancecurves_capacity} displays examples of capacity curves.

\begin{figure}[t]
    \centering
    \includegraphics[width=.4\textwidth,trim=20cm 0 0 0,clip]{fig/performancecurves.pdf}
    \caption{Capacity curve example for a learner class whose complexity can be increased (e.g., a neural network), once for a fixed and finite dataset size and once for the theoretical case of an infinite dataset size.\label{fig:performancecurves_capacity}}
\end{figure}

Capacity curves are interesting from a theoretical viewpoint as they allow us to analyse the intrinsic noise level of the given \emph{data}.
More precisely, one can ask for the performance of a learner of some complexity level on, perhaps, an infinite number of data points.
If this value can be computed for every complexity level, then we obtain a performance curve over the model complexity.
If the number of data points is large enough, this curve can be assumed to be monotonically decreasing.
If we have a maximally flexible learner (such as a neural network) that can, in principle, assimilate any function, then the curve will converge towards the \emph{intrinsic noise} of the data.
That is, no learner can improve over that performance.

\subsubsection{Curriculum Learning}
Curriculum learning is a paradigm inspired by human learning strategies, particularly how humans learn complex tasks by gradually increasing the difficulty of the examples they are exposed to~\citep{wang2022survey}.

In curriculum learning, instead of randomly presenting training instances to the model, instances are presented in a meaningful order, typically from simpler to more complex sets of training instances. This can help the model learn more effectively and converge faster by initially focusing on easier instances that are simpler to learn and gradually introducing more difficult instances.

Learning curves related to curriculum learning come with all sorts of novel challenges, such as interpreting the learning curve in case more complex test instances are provided. 
Therefore, it is hard to make assumptions about what a well-behaved curriculum learning curve would look like. 
For this reason, we consider curriculum learning to be out of scope. 

\section{Modelling a Learning Curve}
\label{sec:curvemodels}
A learning curve \emph{model} is a characterisation of the \emph{true} learning curve derived from an \emph{empirical} learning curve.
The empirical learning curve is the result of sampling from a \emph{stochastic process} that underlies \emph{noise} stemming from randomness in data splits and the learning algorithm itself.
It is typically assumed~\citep{figueroa2012predictingsamplesize,swersky2014freezethawbo,domhan2015speedingup,klein2017fabolas,klein2017lcpredictionwithbnn,mohr2023fast} that, for any learner \learner and any budget $\budget \in \mathbb{N}$, this stochastic process follows the distribution 
\begin{equation}
    \label{eq:noisemodel}
    \lcab \sim \mathcal{N}(\curvemeanab,\curvenoiseab) = \curvemeanab + \mathcal{N}(0,\curvenoiseab),
\end{equation}
where \curvemeanab is either $\learningcurve(\learner, \budget)$ as per Eq.~(\ref{eq:learningcurve:def1}) if modelling a \samplecurve or $\learningcurve(\learner, \sampleanchor, \budget)$ for some (implicit and not further specified) training set size \sampleanchor as per Eq.~(\ref{eq:learningcurve:def2}) when modelling an \itcurve. It assumes a noise that follows a Gaussian distribution with zero mean and dispersion \curvenoiseab that may vary over different anchor sizes.
The assumption of a Gaussian noise is reasonable, because most loss functions form an average over sample-wise scores, which implies a Gaussian distribution through the Central Limit Theorem.
For simplicity, we will not make a difference between the two types of learning curves,  so that anchors are always denoted as budgets \budget (regardless of whether this refers to training set size or iterations, epochs, trees, etc.).

The task of forming a learning curve model for one or multiple learners is inherently one of supervised machine learning and requires the ability to generalise across anchors (and possibly even learners).
In practice, observations are only available for a finite number of learner-anchor combinations $\lcobservationset = \{(\learner_1, \budget_1), (\learner_2,\budget_2), (\learner_3,\budget_3), \ldots \}$, and the task is to learn, for each learner \learner, a model \lcestb that expresses the belief about \lcab for \emph{any} budget \budget, not only those in $\lcobservationset$.
Therefore, it is not enough to simply create an explicit estimate of $\mu_{\learner, \budget_i}$ for the anchors $(a, b_i) \in \lcobservationset$, but some general \emph{pattern} must be learned.
Observe that \learner here is not a parameter of \lcestb since one often does not generalise across learners.
However, some approaches advocate a single model \lcestab that estimates \curvemeanab for any learner-budget \emph{combination}, where $\learner \in \learnerspace$ and $\learnerspace$ is the set of all possible learners \citep{klein2017fabolas,klein2017lcpredictionwithbnn,swersky2014freezethawbo}.

During this process of building a learning curve model, one generally needs to cope with two types of uncertainty.
First, the \emph{aleatoric} uncertainty is \curvenoiseab, which is intrinsic and averaged out in the true learning curve.
Again, this is the uncertainty arising from randomness in the learner itself (if applicable) and random effects in the splits (or more general: data collection) when computing the empirical learning curve.
Second, the \emph{epistemic} uncertainty is the one the learning curve model \emph{itself} has about the estimate of the mean value \curvemeanab.
This uncertainty can be removed by gathering more observations (i.e., extending observation set \lcobservationset).

It is important to understand that epistemic uncertainty generally does not indicate model quality.
Epistemic uncertainty is not related to correctness:
A model can have no epistemic uncertainty (be absolutely sure) about an actually wrong prediction, and similarly, it can be uncertain about a prediction that is actually correct.
Also, epistemic uncertainty gives no indication about whether the \emph{class} from which the predictive model is inferred is suitable for the task, i.e., whether the true curve can be captured by the model that is fitted (e.g., a power law). \citet{hullermeier2021aleatoric} discuss this for the more general case of selecting an appropriate machine learning model.
Recent results suggest that many learning curves are not adequately captured even by a very flexible parametric model, i.e., the 4-parameter MMF model \citep{kielhofer2024learning}, 
which motivates other, possibly non-parametric approaches for modelling, such as the one used in freeze-thaw Bayesian optimisation \citep{swersky2014freezethawbo}.
Therefore, the uncertainty at the meta-level about whether a model \emph{class} is suitable for a task cannot be captured in epistemic uncertainty and must be studied independently.

Modelling uncertainty in learning curve models typically implies modelling the \emph{epistemic uncertainty} about the curve mean \curvemeanab.
This uncertainty refers to \emph{arbitrary} anchors \budget, both those from the observed data $\lcobservationset$ as well as the anchors that were not part of this.
The three common patterns to define beliefs about \curvemeanab are (i)~\emph{point estimates} (no uncertainty is expressed), (ii)~\emph{range estimates}, e.g., confidence intervals for \curvemeanab, or (iii)~\emph{distribution estimates}, which quantify a full belief model over the true value of \curvemeanab.
The aleatoric uncertainty can also be quantified and modelled, e.g., by taking many different samples per anchor.

\begin{figure}[t]
    \centering
    \includegraphics[width=\textwidth]{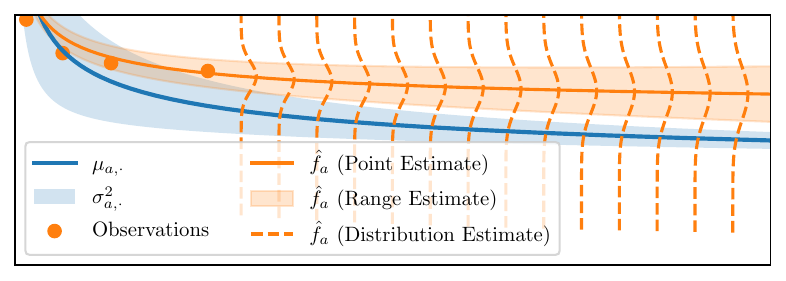}
    \caption{
    Visualisation of various forms of uncertainty in learning curve modelling.
    The blue line $\curvemeana$ represents the true learning curve; the dot $\cdot$ means that it is for all budgets \budget.
    The variance over all possible curves that could be sampled is expressed as $\curvenoisea$.
    The variance stems from the aleatoric uncertainty. 
    Orange points are anchors at which a learner's performance has been observed. 
    From this, an empirical learning curve can be modelled. 
    The orange solid line, area, and dashed lines are point estimate, range estimate, and distribution estimate models, respectively.
    The latter two models also express epistemic uncertainty.}
    \label{fig:modeltypes}
\end{figure}

Fig.~\ref{fig:modeltypes} illustrates all of these concepts together.
The blue line \curvemeana represents the true learning curve as per Eq.~\ref{eq:learningcurve:def1} or Eq.~\ref{eq:learningcurve:def2} for all possible values of \budget.
As this integrates out all possible data splits as well as random factors from the algorithm, each point is an average from a distribution of many possible performance values. 
The variance of this distribution is expressed as $\curvenoisea$.
This variance stems from the aleatoric uncertainty, in the sense that it is high when the aleatoric uncertainty is high, and vice versa.
As the budget $\budget$ (represented on the x-axis) increases, the aleatoric uncertainty typically decreases, and thereby the variance naturally decreases as well. 
The orange elements are related to observations of the learners' performance and the learning curve model.
The orange points (which all together form \lcobservationset) show observations made on the empirical learning curve (as a sample from the blue distributions, this can be one or more per point).
Note that these do not necessarily need to align with \curvemeana or fall within the variance bandwidth. 
The orange solid line shows a point estimate defined by a parametric model obtained from the observations, and in this example, it substantially deviates from the true curve (cf. Sec.~\ref{sec:curvemodels:pointestimates}).
The orange shaded area is a range estimate modelling epistemic uncertainty, which grows as one moves away from the available data (cf. Sec.~\ref{sec:curvemodels:rangeestimates}).
Finally, the dashed orange lines are distribution estimates for different budgets \budget (cf. Sec.~\ref{sec:curvemodels:distributionestimates}). Each of these dashed lines can be seen as a probability density function (rotated 90 degrees) for a certain budget $b$. 
Each curve shown sketches the distribution of the belief about where the true mean may be situated, in this figure modelled through Gaussian distributions.
As we move away from the observations, the shape of these bells grows bigger, indicating higher epistemic uncertainty.

In the following sections, we will explain the three model types in more depth.
We explain the concept for models of the type \lcestb, i.e., models for a specific learner \learner, because this is the most common case.
Generalising a curve model across learners requires additional logic, which we discuss along with the models that utilise such generalised model in Sec.~\ref{sec:literature:anypointandanylearner}.

\subsection{Point Estimates of the Learning Curve}
\label{sec:curvemodels:pointestimates}
The simplest type of learning curve model for a learner \learner just estimates the mean curve values \curvemeanab for any possible budget \budget and ignores uncertainty aspects (solid orange line in Fig.~\ref{fig:modeltypes}).
Given an empirical learning curve in the form of some finite samples from this process at different anchors $B = \{\budget_1,..,\budget_n\}$, a regression model $\lcest_\learner(\cdot\vert\params):\mathbb{N}\rightarrow \mathbb{R}$ is trained with respect to some model class with parameters \params.
A considerable number of different parametric models have been proposed over time for this task.
To our knowledge, the first proposal of such classes was made, apparently independently, by \citet{cortes1993learningcurves} and  \citet{john1996staticvsdynamic} with the three-parametric inverse power law (IPL)
\begin{equation}
    \label{eq:ipl}
    \curvemeanab = \alpha + \beta \budget^{-\gamma},
\end{equation}
where the parameters $\alpha, \beta, \gamma > 0$ need to be optimised to fit the learning curve for learner \learner.
\citet{frey1999powerlawfordecisiontrees} took a simplified variant of that model ($\alpha\budget^{-\beta}$) and compared it to a logarithmic ($\alpha \log \budget + \beta$), and an exponential model ($\alpha \cdot10^{-\beta \budget}$).
While it has been argued, at least for the power-law family, that there is a theoretical foundation for it~\citep{seung1992statistical}, the considered model classes are typically not theoretically motivated but rather pop up in an ad-hoc manner.
For example, \citet{gu2001modelingclassificationperformance} extended the above three classes, without a specific motivation, by a vapor pressure model, the Morgan-Mercer-Flodin (MMF) model, and a Weibull model.

Depending on the purpose of the model, it is essential to distinguish between best-fitting and best-predictive models.
As was pointed out by \citet{gu2001modelingclassificationperformance}, the model class that can best accommodate a given set of anchor points is not always the one that will make the best predictions on a high anchor when having been fit only on some initial anchors.
To understand the learning curve of a learner on a given dataset, one is interested in a best-fitting model class.
For extrapolation, one is interested in a best-predictive one.

Our work does not seek to give a broad overview of different model classes but rather about the usage of such models.
We expose the inverse power law model because it is arguably the most prominent model class and has been advocated by many authors as a good fit for nearest neighbours, SVMs, decision trees, and neural networks~\citep{frey1999powerlawfordecisiontrees,gu2001modelingclassificationperformance,hess2010learning,richter2019approximating}.
However, other models have been proposed, e.g., based on differential equations~\citep{boonyanunta2004predicting} or other physical laws~\citep{gu2001modelingclassificationperformance}, and authors have argued that other models, such as logarithmic shape can be a better fit~\citep{singh2005modeling,gu2001modelingclassificationperformance}.
For an updated and exhaustive overview of used model classes, we refer to the work by~\citet{viering2023theshape}.

\subsection{Range Estimates of the Learning Curves}
\label{sec:curvemodels:rangeestimates}
It has been recognised that incorporating some notion of uncertainty into the model itself is important~\citep{mukherjee2003estimatingdssizerequirements}.
Formally, this amounts to learn a \emph{range estimate} function $\lcest_\learner:\mathbb{N}\rightarrow \mathbb{R}^2$ such that $\lcestb \equiv [u, v]$ with some pre-defined semantic relationship between \curvemeanab and the interval $[u, v]$.
In Fig.~\ref{fig:modeltypes}, this type of estimate is visualised through the orange area.

The semantics of the interval $[u,v]$ depend on what exactly is being modelled, which also implies \emph{how} the model is created.
In the earliest known attempt on this matter, ~\citet{mukherjee2003estimatingdssizerequirements} model the (believed) \emph{interquartile range} of the actual distribution \lcab with this interval, i.e., values that are expected to be observed with a certain probability if sampling at a specific anchor (aleatoric uncertainty).
In this approach, two parametric (i.e., inverse power law) models are built, one from the 25 and one from the 75 quantile for each observed anchor \budget.
Even though the mean does not necessarily lie between these quartiles in general, this is the case in a Gaussian distribution, which is a sensible assumption as explained above.
In contrast, ~\citet{figueroa2012predictingsamplesize} use it to model a confidence interval of \curvemeanab (epistemic uncertainty).
In this specific case, only one parametric model is learned, and the confidence interval around the curve is obtained through analytical rather than stochastic techniques.
The confidence interval-based approach can also be thought of as putting a probabilistic bound on the gap between the predicted performance \lcestb and the true value \curvemeanab.

Again, the mere presence of interval-based predictions that express epistemic uncertainty should not lead to the conclusion or belief that there is a necessary relationship to correctness.
In particular, an epistemic uncertainty of 0 does not imply a correct prediction.
Suppose the model $\lcest_\learner$ is chosen from a class of which the mean curve \curvemeana is not a member. In that case, it is \emph{guaranteed} that there will be wrong predictions, regardless of the epistemic uncertainty expressed by the model.
But even if \curvemeana is among the models from which $\lcest_\learner$ can be built, it can still (and usually will) happen that, based on insufficient observations, a wrong $\lcest_\learner$ will be picked.
In such a case, it is still conceivable that, depending on the probabilistic model on which $\lcest_\learner$ rests, the epistemic uncertainty would be 0 for anchors \budget whereas $\lcab \neq \lcestb$; here \lcestb is just a value since the epistemic uncertainty of 0 means that the interval \lcestb has only one value.

\subsection{Distribution Estimates of Learning Curves}
\label{sec:curvemodels:distributionestimates}
In a more ambitious case, we can try to learn a full \emph{belief model} of the learning curve  \learningcurve.
Formally, this amounts to learn a \emph{distribution estimate} function 
$\lcest_\learner:\mathbb{N}\rightarrow \{p~\vert~p\text{ is a distribution in the domain of the performance measure}\}$.
Fig.~\ref{fig:modeltypes}, this model corresponds to the sequence of dashed orange distributions (in this figure, displayed for only 14 values of budget \budget).

As with the previous approaches, one typically uses parametric functions (e.g., the inverse power law) as a basis but specifies \emph{distributions over their parameters} instead of a single (based on the maximum likelihood) assignment.
The distribution over parameters then induces a distribution of the space of learning curves.
Such a belief model is, for example, well-defined in a Bayesian framework that defines the posterior distribution of models given the observed data and assumes a certain model class.
This posterior distribution cannot be efficiently computed exactly but approximate it through sampling~\citep{domhan2015speedingup,klein2017lcpredictionwithbnn}.

Clearly, distribution estimates are the most flexible way of modelling uncertainty and allow many interesting operations.
In particular, one can quantify the probability that the limit performance of a learning curve will be above or below some threshold $\tau$, which is very useful for confidence-based early discarding \citep{domhan2015speedingup}.

\section{A Framework to Categorise Learning Curves Methods for Decision Making}
\label{sec:taxonomy}
Based on the common ground of learning curves and their models introduced in Sec.~\ref{sec:background} and Sec.~\ref{sec:curvemodels}, this section presents a framework for \emph{categorising} decision-making methods that use learning curves.
We identify three orthogonal criteria along which those approaches can be categorised.
The first criterion relates to the \emph{decision-making situation} in which learning curves are used.
We discuss these situations exhaustively in Sec.~\ref{sec:taxonomy:situations}.
The second dimension covers the \emph{technical question} that is answered about a learning curve to support the decision.
For example, are we interested in the saturation point or a complete model?
These technical questions are sketched in Sec.~\ref{sec:taxonomy:technicalquestions}, and we structure the literature review of Sec.~\ref{sec:literature} according to this axis.
Finally, different \emph{data resources} can be used to conduct an analysis with learning curves, e.g., other learning curves or features describing the datasets or the learning algorithms.
These resources are covered in Sec.~\ref{sec:taxonomy:resources}.

\subsection{Types of Decision-making Situations}
\label{sec:taxonomy:situations}
Learning curves are an important resource in at least three types of decision-making situations:
\begin{enumerate}
    \item \emph{Quantitative Data Acquisition}
    Consider the situation where a data scientist has a model trained on a set of observations, the performance is known, and there is the option to spend additional resources (e.g., money or labelling effort) to obtain additional training observations. The decision that needs to be made is:
    The acquisition of how many more labels is economically reasonable?
    This question has an obvious connection to the field of \emph{active learning}, which addresses the question of \emph{which} instances should be labelled next (qualitative acquisition).
    Another question is whether we should acquire other features instead.
    
    \item \emph{Early Stopping} (of training an independently considered model).
    In the situation where one is committed to some specific learner (a learning algorithm \emph{and} its hyperparameters), minimising the training effort is a reasonable goal.
    Specifically, if large amounts of data are available and training is costly, the aim is to train until the saturation point is reached.
    Being able to detect or predict whether a learner's performance saturates after a given number of observations or iterations can support making decisions on this.
    
    \item \emph{Early Discarding} (in model selection).
    Similarly, if we want to \emph{select} from various models, we want to stop the evaluation of a candidate when we are sure that it is not \emph{competitive} to the current best solution.
    We compare the learner performance to that of \emph{another} learner instead of its own performance on more training investment.
    For example, consider the situation where the learning curve of an algorithm seems to approach the saturation point, and we have already seen a superior model before, of which it is unlikely that the current algorithm will improve over. In this case, we can discard the performance of this learner based on the performance in relation to other models.
\end{enumerate}
There is a large methodological overlap in creating a decision basis among all these decision-making situations.
For example, whether more data points would be helpful to improve performance is related to the question of the training size that should be chosen to minimise training effort.
Both questions, at their core, ask for the saturation point of the learning curve.

In the following sections, we will discuss each of the three decision-making situations in more depth.

\subsubsection{Quantitative Data Acquisition}
\label{sec:acquisition}
Quantitative data acquisition focuses on the question of how \emph{many} training examples should be considered, given that they are all sampled i.i.d. from the same source.
Quantitative data acquisition does not consider or pay attention to the possibility of acquiring \emph{specific} instances, which would be considered in qualitative data acquisition.
Qualitative data acquisition is mainly studied in the field of \emph{active learning} and does not ask whether or how many instances should be acquired but for \emph{which} instances a label should be acquired.
Since active learning undermines the i.i.d. assumption, it generates a different type of learning curve and is not covered in this survey.

The relevance of learning curves for quantitative data acquisition rises from their ability to give insights into intrinsic properties of the data \emph{source} as well as into the relationship between the number of training examples and the \emph{utility} of having that number of samples.
On the one hand, intrinsic properties refer to the intrinsic noise of the data~\citep{cortes1994limits}, which tells us about the best possible performance of any learner no matter how much data would be available from the source.
If we know that, with the given data, we already achieve a performance close to the intrinsic noise, then data acquisition should focus on acquiring additional \emph{features} instead of new instances.
On the other hand, the utility is mainly determined by the cost of acquiring (additional) examples, the performance obtained with a certain number of samples, and the cost to train a model with a given number of instances~\citep{last2007predicting,weiss08maximizingclassifierutility,last2009improving}.
In this economic context, there are mainly five questions that can be considered:
\begin{enumerate}
    \item \emph{Possibility.} Can the classification performance be improved by more data?
    
    \item \emph{Potential.} What is the best possible predictive performance given unlimited training observations?
    
    \item \emph{Maximization Principle.} By how much can the predictive performance be improved if there is a budget for a fixed number of additional data points?
    
    \item \emph{Minimization Principle.} How many instances are necessary to obtain a certain degree of predictive performance?
    
    \item \emph{Utility maximization.} Which sample size maximises a given utility function?
\end{enumerate}
In the context of data acquisition, we typically deal with \samplecurves. 
Furthermore, one is often not committed to a particular learner; therefore, the performance measure in the above questions is implicitly the best one of a \emph{portfolio} of learners.
That is, one assumes a set of learners that is considered admissible for the prediction task due to external restrictions.
In general, due to the ability to parameterise learners, this set is usually infinite.
When referring to the portfolio's performance at a specific anchor point, we are interested in its best-performing algorithm at that specific point at the learning curve~\citep{mohr2023fast}.
Of course, if one is committed to one particular learner, then the situation simplifies to a portfolio of size 1.

Since data acquisition is not for free, it is sensible to relate potential predictive performance improvements with the costs to collect the additional labels.
Therefore, instead of looking only at predictive performance, one looks at \emph{utility} of an anchor point.
While performance typically only improves with an increased number of observations, the utility also considers acquisition costs, which negatively affect the utility.
Therefore, the goal is to decide how many instances should be labelled to maximise utility, i.e., how many additional instances are justified before the added value no longer outweighs the additional costs~\citep{last2007predicting,weiss08maximizingclassifierutility}.

\subsubsection{Early Stopping}

\begin{figure}[t]
    \centering
    \includegraphics[width=\textwidth]{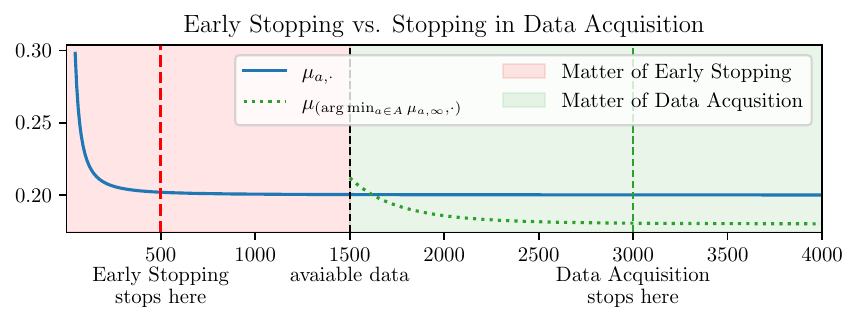}
    \caption{Early stopping with \samplecurves stops the training process of a single learner \learner at its saturation point using its (actually empirical) learning curve.
    It shows the learning curve of two learners, the blue curve and the green dashed curve. 
    The latter has the best saturation performance. 
    The available data marks a restriction to this process.
    In contrast, data acquisition considers the available data as the decision variable, and it stops collecting data when \emph{all} learners have reached their saturation performance.
    }
    \label{fig:earlystoppingvsdataacquisition}
\end{figure}

Early stopping means interrupting the training process of a learner if the learning curve has converged.
The term `early' refers to the fact that the learning process would normally be continued, e.g., because more data is available or other stopping criteria are not yet satisfied.
That is, one uses the learning curve to judge that, despite more available resources or other criteria that would encourage further training, investing more time will not improve the performance of the considered model class any further.
The training process can then be stopped early, i.e., earlier than if that criterion would not be used.
Fig.~\ref{fig:earlystoppingvsdataacquisition} shows this logic in the left (red) part in which the blue learning curve of learner \learner is used to detect that not all the data is necessary and only 500 training instances are used (to save training time).

Early stopping can be applied to both \samplecurves and \itcurves. 
Early stopping in \samplecurves means retraining a model on different training set sizes to create an empirical learning curve with training set size as the budget.
This can make sense if we do not already know that the saturation point is larger than the available dataset size; otherwise we should immediately train on the complete dataset.
We can then try to analyse the \samplecurve of the learner for increasing training sizes and stop as soon as we find that performances do not change significantly between two anchors~\citep{john1996staticvsdynamic,provost1999efficientprogressivesampling}.
For iterative learners (such as neural networks), an \itcurve is usually a by-product that can cheaply be created in parallel to learning.
Hence it might seem more appropriate to do early stopping based on an \itcurve rather than the \samplecurve.
An additional advantage of early stopping in \itcurves is that it can help avoid over-fitting, e.g., in neural networks \citep{bishop1995regularization,goodfellow2016deep} or gradient boosting.
While it is conceivable that building \samplecurves, even for iterative learners, could be useful in some cases, we are not aware of any such work being done for early stopping (or any other purpose).

The early stopping problem can be addressed retrospectively and projectively.
Retrospective early stopping means to stop after observing the saturation point.
Projective early stopping means to \emph{predict} the saturation point before it is reached and stop precisely at the (believed) saturation point.
The projective approach is particularly important in the case of \samplecurves.

Early stopping in \samplecurves and data acquisition might seem similar since both define a sample size at which a process should be stopped.
However, the concepts are fundamentally different in three ways:

\begin{enumerate}
    \item \emph{Stopped Process:} Early stopping means to stop a \emph{training} process (at the saturation point) run in a \emph{machine}. In contrast, the decision-making situation in data acquisition is to stop the \emph{data acquisition} process (at the economic saturation point). The latter is sometimes carried out by \emph{humans}.
    
    \item \emph{Role of Available Data:} In early stopping, the available amount of training data is a given constraint under which early stopping operates, and data acquisition precisely seeks to control this quantity in an economically optimal fashion.

    \item \emph{Used Performance Curve:} Early Stopping uses a single learning curve of learner \learner to decide upon early stopping of the training of \learner.
    Data acquisition uses the learning curves of all learners under consideration (i.e., a finite set $A$) and considers the budget-wise best performance achievable (by \emph{any} learner).
\end{enumerate}
To clarify this difference, consider also the right (green) part of Fig.~\ref{fig:earlystoppingvsdataacquisition}.
The learner \learner may be the best solution available given the 1500 data points (in this case, \learner barely needs 500 of them to attain saturation performance).
However, if more data \emph{were} available, then at least one other learner could take advantage of that additional data and outperform \learner.
The figure shows the curve of an optimal learner $a^* \in \arg\min_{\learner \in A} \mu_{\learner,\infty}$ that has the best performance if no limit is posed on the available training data (as in \citet{cortes1994limits}).
Only after 3000 instances, \emph{no} learner will improve the overall possible performance anymore; therefore, at this point, the data acquisition process stops.

\subsubsection{Early Discarding}
\label{sec:decisionsituations:earlydiscarding}
In many setups, the learner itself is a matter of optimisation.
Consider the situation where we have a (possibly infinite) \emph{set} of learners, e.g., a finite set of algorithms, each of which can be instantiated with a possibly infinite number of hyper-parametrisations.
The task is to find the (hyper-parametrised) learner which performs best for the given data of size \sampleanchor in the sense that it creates, on average, the best model.
Formally, if \learnerspace is the (infinite) set of parametrised learners, the goal is to find

\begin{equation}
    \label{eq:earlydiscarding}
    \arg\min_{\learner \in A} ~ \learningcurve(\learner, \sampleanchor).
\end{equation}
This task is commonly known as \emph{model selection}.

While it is uncommon in literature to be so explicit and describe the model selection problem through the value of the learning curve at some sample size, this formulation is rather precise and insightful.
It emphasises that which learner is best might depend on the number of available training points.
Note that one needs to separate some portion of the data for validation in practice to \emph{estimate} model performances.
In other words, most approaches in practice do not even address the above problem but instead
\begin{equation}
    \label{eq:earlydiscarding:validationreduced}
    \arg\min_{\learner \in A} ~ \learningcurve(\learner, \ceil{\alpha \sampleanchor}),
\end{equation}
where $\alpha \in ]0,1[$ (open interval) is the training portion, typically between 70\% and 90\%, where the remaining portion of $1-\alpha$ is used to estimate $\learningcurve(\learner, \ceil{\alpha \sampleanchor})$, typically in some (possibly repeated) hold-out validation.

We could conceive that this procedure of estimating $\learningcurve(\learner, \ceil{\alpha \sampleanchor})$ might involve the construction of an empirical learning curve as a sub-routine or on-the-fly.
First, if \learner is an iterative learner, then the model performance $\learningcurve(\learner, \ceil{\alpha \sampleanchor}) = \learningcurve(\learner, \ceil{\alpha \sampleanchor}, \timeanchor^*)$ is the performance of the \itcurve at some point $\timeanchor^*$ where the learning process is stopped.
Therefore, the whole \itcurve $\learningcurve(\learner, \ceil{\alpha \sampleanchor}, \timeanchor^*)$ is available for analysis.
Second, even if \learner is not incremental, one could create an schedule $\{\alpha_1,..,\alpha_k\}$ of increasing $\alpha_i \leq \alpha$ and thereby create a \samplecurve \citep{mohr2023fast}.
Such a \samplecurve would also offer the perspective, via extrapolation, to address Eq.~(\ref{eq:earlydiscarding}) rather than just Eq.~(\ref{eq:earlydiscarding:validationreduced}).

In the light of the availability of such a (partial) empirical learning curve, \emph{early discarding} is the practice of aborting the performance estimation procedure of a candidate as soon as it becomes apparent from that curve that the candidate cannot be the solution to the above optimisation problem.

Formally, this is to drop a candidate \learner as soon as the criterion
\begin{equation}
    \label{eq:early_discarding}
    \learningcurve(\learner, \refpoint) > \min_{\learner^* \in \learnerspace}\learningcurve(\learner^*,\refpoint)
\end{equation}
can be verified, where \refpoint is usually \sampleanchor or $\ceil{\alpha \sampleanchor}$.
In other words, as soon as it can be shown that $\learner$ is not the best learner of all possible learners $\learnerspace$, no further resources should be committed to training learner $\learner$.
Note that, even though the terms are frequently mixed up in literature, this is very different from early stopping, in which the convergence of the curve of a single learner is considered in isolation (cf. Fig.~ \ref{fig:decision_situations}).

Early discarding has been applied to both observation~\citep{mohr2023fast,mohr2021towards,ruhkopf2022masif,adriaensen2023efficient} and iteration~\citep{swersky2014freezethawbo,domhan2015speedingup,klein2017lcpredictionwithbnn,ruhkopf2022masif,adriaensen2023efficient} learning curves.
In the first case, one is sampling from $\learningcurve(\learner, \cdot)$ at different anchors \sampleanchor and hopes to be able to drop sub-optimal candidates at $\sampleanchor \ll \refpoint$ anchors (much) smaller than the target size \refpoint.
In the second case, one always uses the complete dataset (or at least all data designated for training, say \sampleanchor), observes samples of $\learningcurve(\learner, \sampleanchor, \cdot)$ for different anchors (maybe epochs) \timeanchor, and seeks to avoid convergence if it can be foreseen that the convergence performance will be sub-optimal.

Early discarding is more aggressive than early stopping because it does not need to wait until the learning curve converges.
On the contrary, one tries to avoid reaching convergence since this is considered a waste of resources in the case that the learner performs sub-optimal.
In an extreme case and depending on available knowledge about learners (cf. Sec.~\ref{sec:taxonomy:resources}), one could only use a single point of an empirical learning curve to discard a candidate.
    
Situations in which early discarding plays a role can be further classified into horizontal and vertical scenarios (and a mixture of the two):
\begin{enumerate}
    \item \emph{Horizontal Model Selection.}
    Horizontal model selection implies an apriori fixed \emph{finite} set of learning algorithms, from which one has to be selected. Empirical learning curves are grown iteratively for the whole set or shrinking subsets of it.
    Successive halving and related works are a prominent example of horizontal decision making~\citep{vandenbosch2004wrappedprogressivesampling,jamieson2016nonstochasticbestarm}.
    However, these approaches only consider the last anchor point (rather than the complete learning curve).
    
    \item \emph{Vertical Model Selection.}
    Vertical means that the set of learners is generally not limited to a finite set, and the set of \emph{evaluated} learner candidates evolves over time (i.e., not fixed apriori).
    Learners are evaluated one \emph{after} another.
    Each learner is evaluated in an iterative fashion to grow a learning curve and allow for early discarding.
    Examples are the early discarding routine for deep networks by \citet{domhan2015speedingup} or, more generally, for learning curve cross-validation~\citep{mohr2023fast,mohr2021towards}.
    
    \item \emph{Diagonal Model Selection.}
    This case is similar to the vertical decision-making situation with the difference that one does allow to \emph{continue} the evaluation of a candidate at a later point.
    Hence, candidates are not evaluated one after another, but the evaluation of different candidates can be interleaved.
    Examples are Bayesian optimisation-based approaches to pause and continue evaluations of (not necessarily iterative) learners~\citep{swersky2014freezethawbo,klein2017fabolas}.
    Non-iterative learners must be trained from scratch with the increased budget.
    
    Another approach that addresses this type of decision-making situation is Hyperband~\citep{li2017hyperband} and Bayesian optimisation based on progressive sampling~\citep{zeng2017progressivesampling}.
    However, neither of these approaches considers learning curves even though they implicitly construct them.
    Decisions are taken based on the observations of the largest anchor point considered so far.
\end{enumerate}

\subsection{Technical Questions Asked About Learning Curves}
\label{sec:taxonomy:technicalquestions}

\begin{figure}[t]
    \centering
    \includegraphics[width=\columnwidth]{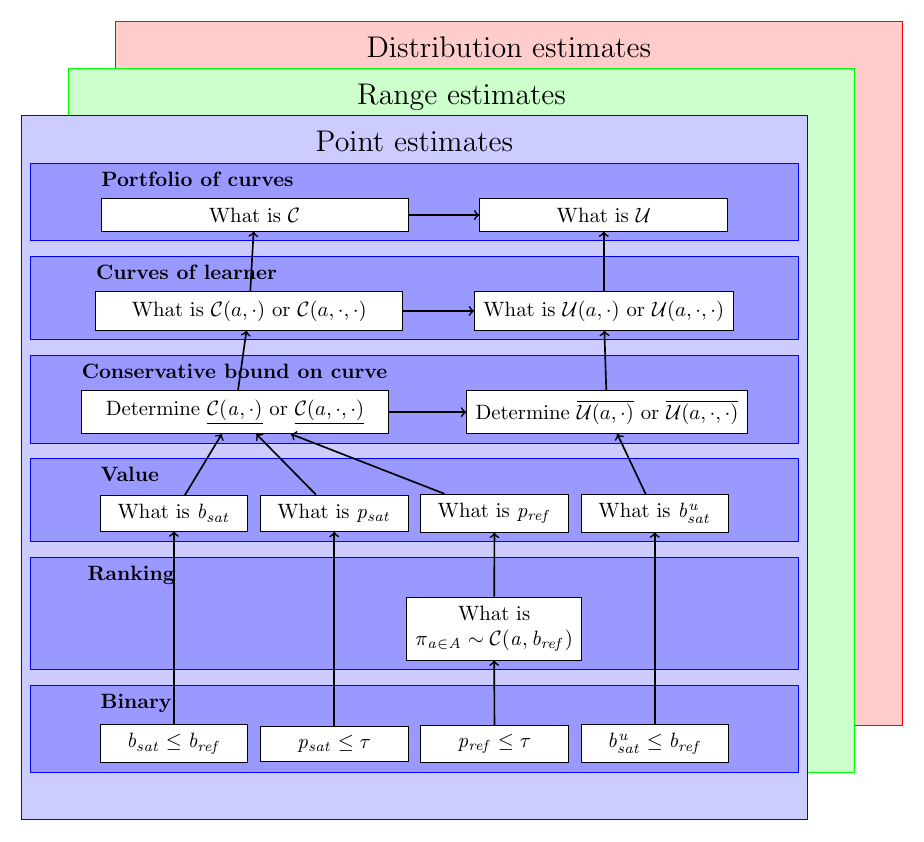}
    \caption{Technical questions that can be asked on learning curves.}
    \label{fig:questions:learningcurve}
\end{figure}

A plethora of questions can be asked about learning curves.
Fig.~\ref{fig:questions:learningcurve} gives an overview of these questions.
The figure is organised in three layers (depth dimension) corresponding to the three types of estimates discussed in Sec.~\ref{sec:curvemodels}.
Each layer consists of a set of questions that can be posed about learning curves.
From bottom to top, the questions are ordered by complexity, and an arrow from one question to another indicates that the question with the incoming arrow is more general. Answering the more general question also implies answering the less general question.

In the simplest case, we can answer a binary question.
There are four relevant questions, i.e., (i)~whether some specific anchor point, e.g., the dataset size, is beyond the saturation point ($\satpoint \le \refpoint$), (ii)~whether the performance of a learner at the saturation point is better than some baseline $\tau$ ($\satperformance \le \tau$), (iii)~whether the performance $\refperformance := \learningcurve(\learner, \refpoint)$ at some reference point \refpoint is better than some threshold $\tau$, or (iv) whether a specific anchor point is beyond the \emph{utility-based} stopping point ($\economicstoppingpoint < \refpoint$).
To our knowledge, the only approaches in this category are those implicitly answering question (iii)~by discarding candidates that are not believed to be competitive \citep[see, e.g.,][]{petrak2000fast,jamieson2016nonstochasticbestarm,zeng2017progressivesampling}; here $\tau = \min_{\learner \in A}\learningcurve(\learner, \refpoint)$ is the (unknown) best performance of any learner on the target size.

A family of slightly more general questions tries to \emph{order} a set \learnerspace of learning algorithms w.r.t. their performance at some (future) anchor \refpoint.
We denote this ordering as \candidateranking.
Here, \refpoint is typically the maximum available training data, even if an \itcurve is considered because then this is the termination performance of that curve.
In the simplest case, we could ask for a concrete pair of two learning algorithms $\learner_1$, $\learner_2$ whether $\learningcurve(\learner_1,\refpoint) \geq \learningcurve(\learner_2,\refpoint)$, i.e., which will perform better at some reference point.
For example, the work by \citet{leite2005predictingrelativeperformance} answers this question.
If this question is simultaneously asked for a set of or even all possible pairs of algorithms, one asks for a partial or even the full ranking \candidateranking of algorithms.
A particular case is to ask only for the best algorithm $\learner^* = \arg\min_\learner \learningcurve(\learner,\refpoint)$, which implicitly answers that $\learningcurve(\learner^*,\refpoint) \leq \learningcurve(\learner,\refpoint)$ for any learner \learner but without explicitly asking for any other comparisons.
Still, the comparison is merely qualitative, and answering this question does not necessarily require to quantify any aspect of any learning curve.

The above questions are merely qualitative and not quantitative, which gives rise to a third level of complexity, where the concrete values of \satpoint~\citep{provost1999efficientprogressivesampling}, \satperformance~\citep{cortes1993learningcurves}, \refperformance~\citep{leite2003improvingprogressivesampling,leite2004improvingprogressivesampling,chandrashekaran2017speedinguphpo,baker2018acceleratingnas}, or \economicstoppingpoint~\citep{weiss08maximizingclassifierutility} are being modelled.
Here, the reference performance \refperformance is the performance at a fixed reference point, often the number of training samples available for cross-validation.

All questions up to this point produce closed answers in the sense that the answer is either a boolean value, a number, or a finite ranking of candidates.
At the fourth level, the task is to make assertions about arbitrary points of a learning curve.
However, the answers do not yet refer to the value of the learning curve itself but only a \emph{bound} on those values.

At a fifth level, we could eventually ask for a model of the whole learning curve of a learner \learner, i.e., $\learningcurve(\learner, \cdot)$ for \samplecurves~\citep{john1996staticvsdynamic,frey1999powerlawfordecisiontrees,gu2001modelingclassificationperformance,mukherjee2003estimatingdssizerequirements,figueroa2012predictingsamplesize} or $\learningcurve(\learner, \sampleanchor, \cdot)$ for \itcurves given a fixed (sample) anchor \sampleanchor~\citep{cortes1993learningcurves,domhan2015speedingup}.
We can ask a similar question for the utility curve. While the question is on the same level, it is more general, as it is \emph{based} on the learning curve itself~\citep{last2007predicting,last2009improving,weiss2006maximizingclassifierutility} and combines it with other information such as acquisition costs.

Finally, at the sixth and most general level, we could ask for a model of the whole performance function \learningcurve, i.e., the model of the learning curves across all learners~\citep{swersky2014freezethawbo,klein2017fabolas,klein2017lcpredictionwithbnn}.
Analogously, this question could be asked for the whole utility function \utilitycurve, which is arguably the most complex and general question that can be asked, although we are not aware of any works that have done so.

Since all of the above questions are answered based on observational statistics, we can also consider noise and uncertainty aspects for the quantitative questions.
In the simplest case (blue layer, cf. Sec.~\ref{sec:curvemodels:pointestimates}), we only get point estimates, i.e., the estimate of \learningcurve at one or a set of points.
Since these estimates are always afflicted with uncertainty, it is reasonable to ask for quantifications of this uncertainty.
One form is to express \emph{strict bounds} as in the fourth layer, which may be derived from assumptions about the learning curve shape, e.g., convexity \citep{mohr2021towards,mohr2023fast}.
Another form is to express \emph{probabilistic bounds} like confidence intervals around \learningcurve~\citep{figueroa2012predictingsamplesize,koshute2021recommending} (cf. Sec.~\ref{sec:curvemodels:rangeestimates}).
In the most general form, we can ask for a full \emph{belief model} of the learning curves, specifying a probability distribution over the values of \learningcurve at one point or a set thereof~\citep{swersky2014freezethawbo,domhan2015speedingup,klein2017fabolas,klein2017lcpredictionwithbnn} (cf. Sec.~\ref{sec:curvemodels:distributionestimates}).

It is common practice to solve relatively simple questions by implicitly answering more complex ones.
For example, a typical question in the context of model selection is whether the performance of a candidate learner at some given data or in the limit will beat a known baseline~\citep{leite2005predictingrelativeperformance,vanrijn2015fastalgorithmselection,domhan2015speedingup}.
This is the binary question of $\refperformance \le p^*$, where $p^*$ is the best-known performance.
Often, this question is answered by \emph{estimating} \refperformance (maybe \limperformance) explicitly and then comparing it to $p^*$~\citep{leite2005predictingrelativeperformance,vanrijn2015fastalgorithmselection}, which is an answer to a slightly more complicated question.
\citet{swersky2014freezethawbo,domhan2015speedingup} built an explicit curve model, for estimating \limperformance and \refperformance, respectively.
The approach answers this binary question by building an entire learning curve model and then derives the binary answer from it.
The rationale behind this is the notion that one often needs rather complex models to find a high-quality answer to a simple question, and it is just a side effect that one can then even answer other questions with those models.

\subsection{Used Data Resources for Inference}
\label{sec:taxonomy:resources}
Above we have discussed a series of questions that can be asked about learning curve properties, which in turn are important for decision making in a specific context.
Of course, answering these questions requires specific informative resources.
In a concrete decision-making situation, we are typically confronted with a dataset and a learner or a portfolio of learners.
We call this the \emph{target} dataset and the \emph{current} learner.
That is, we want to say something about the learning curve of the current learner in a domain in which we have a finite (the target) dataset \dataset available.

Fig.~\ref{fig:resources} shows the types of data resources that can be used to answer questions about learning curves.
We can utilise empirical learning curves gathered on the target dataset, empirical learning curves gathered on other datasets, dataset meta-features, and features describing the learners.
Learning curves gathered on the target dataset come at a particular computational cost, as they need to be generated during the process. 
Typically, when modelling the current learner on the target dataset, a \emph{partial} empirical learning curve is constructed, which can then be step-wise extended or discarded~\citep{provost1999efficientprogressivesampling,leite2003improvingprogressivesampling,domhan2015speedingup}.
When using learning curves of other datasets, those are usually available up to a large portion of the dataset size (with a specific schedule of anchors) in the case of \samplecurves or until convergence in the case of \itcurves.
This is because these curves could be prepared offline before the target dataset became available~\citep{leite2003improvingprogressivesampling,leite2010activetesting,vanrijn2015fastalgorithmselection}.
Both learning curves of the current learner and other learners can be utilised for this.
In the context of model selection, various learners are usually evaluated. Therefore we can acquire various learning curves of other learners on the target dataset~\citep{chandrashekaran2017speedinguphpo,swersky2014freezethawbo,klein2017fabolas,klein2017lcpredictionwithbnn,baker2018acceleratingnas}.

\begin{figure}[t]
\centering
\begin{tikzpicture}[node distance=1cm, auto]  
\tikzset{
    mynode/.style={rectangle,rounded corners,draw=black, top color=white, bottom color=yellow!20,thick, inner sep=1em, minimum width=6em, minimum height=4.5em, align=center, font=\footnotesize, blur shadow={shadow blur steps=5}},
    myarrow/.style={-, >=latex', thick}
}  
\node[mynode] (resources) {Data\\Resources};  
\node[below=1.5cm of resources] (dummylvl1) {}; 
\node[mynode, left=2.25cm of dummylvl1] (metaf) {Dataset\\Meta-\\Features};  
\node[mynode, right=2.25 cm of dummylvl1] (learnerf) {Learner\\Features};

\node[mynode, left=0.01 cm of dummylvl1] (actived) {Target\\Dataset};  
\node[mynode, right=0.01 cm of dummylvl1] (otherd) {Other\\Datasets};

\node[below=2.0cm of dummylvl1] (dummylvl2) {}; 
\node[mynode, left=2.25 cm of dummylvl2] (adac) {Curves on\\Current\\Learner};
\node[mynode, left=0.01 cm of dummylvl2] (adoc) {Curves on\\Other\\Learners};
\node[mynode, right=2.25 cm of dummylvl2] (odac) {Curves on\\Current\\Learner};
\node[mynode, right=0.01 cm of dummylvl2] (odoc) {Curves on\\Other\\Learners};

\draw[myarrow] (resources.south) -- (metaf.north);	
\draw[myarrow] (resources.south) -- (learnerf.north);
\draw[myarrow] (resources.south) -- (actived.north);	
\draw[myarrow] (resources.south) -- (otherd.north);	

\draw[myarrow] (actived.south) -- (adac.north);	
\draw[myarrow] (actived.south) -- (adoc.north);
\draw[myarrow] (otherd.south) -- (odac.north);	
\draw[myarrow] (otherd.south) -- (odoc.north);

\end{tikzpicture}
\caption{Taxonomy of data resources for learning curve analysis \label{fig:resources}}
\end{figure}

Other types of data resources that can be used are meta-features on the datasets and learner features.
These are measurable qualities of the dataset and learner, respectively, and these can indicate how similar specific datasets (or learners) are.
These give the decision-making algorithm a sense of which learning curves are more informative for the current learner and target dataset. 
To the best of our knowledge, the only line of research utilising meta-features for learning curve modelling is the work of \citet{leite2008selecting,leite2010activetesting,ruhkopf2022masif}.
The description of learners through features for the sake of model prediction is specifically prevalent in the analysis of \itcurves~\citep{swersky2014freezethawbo,klein2017lcpredictionwithbnn,baker2018acceleratingnas}.
The development and analysis of meta-features is a research field in its own right; for more information, we refer the reader to~\citet{brazdil2022metalearning}.

\section{Literature Review on learning curve extrapolation methods}
\label{sec:literature}
This section presents the literature review that covers methods to model and utilise learning curves.
We organise approaches based on the key problem they resolve on a rather abstract level and independent of the purpose or type of decision-making situation in which they were presented.
We organise it along the framework presented in Fig.~\ref{fig:questions:learningcurve}, particularly along the axis that reflects the technical question asked about a learning curve.
The motivation to not use the type of decision-making situation, which is also a prominent property of these methods, is that the same approach can be used in \emph{different} decision-making situations.
For example, \citet{leite2004improvingprogressivesampling} present and motivate an approach to identify the portion of some given data that should be used for training (i.e., determine the saturation point), but the same approach could be used to determine how much more data would be needed to obtain saturation performance.
Similarly, \citet{domhan2015speedingup} present an approach that aims to decide during training whether a neural network will become competitive (ask for saturation performance); however, they did this by modelling the complete learning curves.

\begin{figure}
  \begin{center}
    \includegraphics[width=\textwidth]{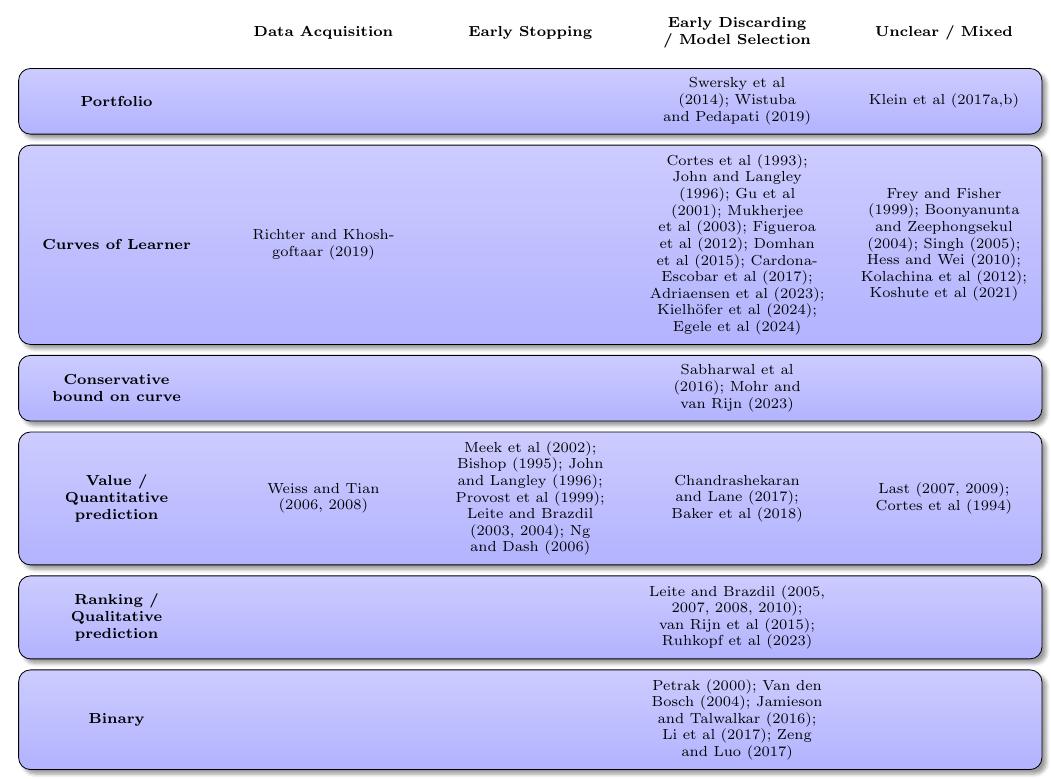}
    \caption{Overview of the methods (indicated by the bibliographic reference) covered and categorised in this framework. 
    Each method is categorised along the problem type they solve (vertical axis) and the type of decision-making situation they explicitly address (horizontal axis). 
    Some methods are employed to address multiple decision-making situations. 
    Some papers do not explicitly state which decision-making situation is being addressed, or address multiple. 
    These are categorised in the last column.
    Several models can be used for more decision-making situations than the original paper evaluated them on. For example, the portfolio approaches could in theory be used for any decision-making situation. 
    \label{fig:problem_solution}}
  \end{center}
\end{figure}

Fig.~\ref{fig:problem_solution} gives an overview of all the methods categorised in this framework. 
We make the following two observations.
\begin{itemize}
\item Most learning curve methods address the early discarding / model selection decision-making situation. 
This implies that there is an opportunity for more research on, for example, data acquisition or early stopping. 
We note that research towards active learning provides many approaches that handle data acquisition (which we do not cover), which might serve as a basis for a literature search. 
\item Second, most methods are centred around the middle levels of problem complexity they address, i.e., predicting the actual value of a learner at a certain point and predicting the complete curves of a learner. 
It seems logical that there are benefits for exploiting the situation where either the complete portfolio is modelled (e.g., the benefit of parameter sharing, the opportunity of model acquisition) or the binary problem is solved (because of the simplicity of the problem definition). 
Methods addressing the binary situation, such as Successive Halving and Hyperband, have attracted quite some attention. 
\end{itemize}
To further structure the overview, we divide the whole literature on methods for learning curves into two roughly even groups according to the \emph{usage} of a learning curve model as described in Sec.~\ref{sec:curvemodels}.
Approaches in the first group do not employ a learning curve model. These approaches address the questions defined in the four lowest levels of the framework (see Sec.~\ref{sec:literature:nomodel}).
In contrast, approaches of the second group, i.e., which employ a learning curve model, address questions in the top two levels of the framework (see Sec.~\ref{sec:literature:withmodel}).
Approaches with a curve model are more general, but that does not necessarily mean they give better answers to simpler questions.
In fact, \citet{kielhofer2024learning} show that the model-free MDS approach discussed in Sec.~\ref{sec:literature:ranking} in specific situations outperforms a parametric model such as the ones presented by \citet{gu2001modelingclassificationperformance} discussed in Sec.~\ref{sec:literature:anypoint:point}.
In total, the approaches cover 10 of the questions discussed in Sec.~\ref{sec:taxonomy:technicalquestions}, which are organized in the following sub-sections:

\begin{enumerate}[5.1.]
    \item Approaches without learning curve model:

    \begin{enumerate}[1.]
        \item Is the target performance of a learner worse than the one of the best learner, i.e., $\refperformance > \min_{\learner \in A}\learningcurve(a, \refpoint)$?
    
        \item What is the ordering \candidateranking of the learning algorithms w.r.t. their performance at some target anchor?
        
        \item What is the saturation performance (\satperformance)?
    
        \item What is the saturation point (\satpoint)?
        
        \item What is the utility-based stopping point (\economicstoppingpoint)?
        
        \item What is the value of the learning curve at a specific (fixed and known) point? ($\learningcurve(\learner,\sampleanchor)$ or $\learningcurve(\learner,\sampleanchor,\timeanchor)$)

        \item What is a lower/upper bound of the learning curve at any point? ($\underline{\learningcurve(\learner, \cdot)}$ or $\underline{\learningcurve(\learner, \sampleanchor, \cdot)}$)
    \end{enumerate}

    \item Approaches with a model for the learning curve:

    \begin{enumerate}[1.]
        \item What is the value of the learning curve at \emph{any} (queryable) point? ($\learningcurve(\learner,\cdot)$ or $\learningcurve(\learner,\sampleanchor, \cdot)$)

        \item What is the \emph{utility} at an arbitrary anchor point? ($\utilitycurve(\learner,\cdot)$ or $\utilitycurve(\learner,\sampleanchor, \cdot)$)
        
        \item What is the value of the learning curve at any queryable point of \emph{any} queryable learner? ($\learningcurve(\cdot,\cdot)$)
    \end{enumerate}
 \end{enumerate}
We organise the rest of this section exactly according to this scheme.

For each approach, we always consider the most general problem it solves, independently of how this solution is used in the context of a paper.
For example, \citet{domhan2015speedingup} decide whether the saturation performance of a learner beats some threshold (question at the binary level) but develop a learning curve model and are hence discussed alongside the approaches for question 1 in Sec.~\ref{sec:literature:performanceatanypoint}.

\subsection{Approaches Without Learning Curve Models}
\label{sec:literature:nomodel}
\begin{figure}[t]
    \centering
    \includegraphics[width=\textwidth]{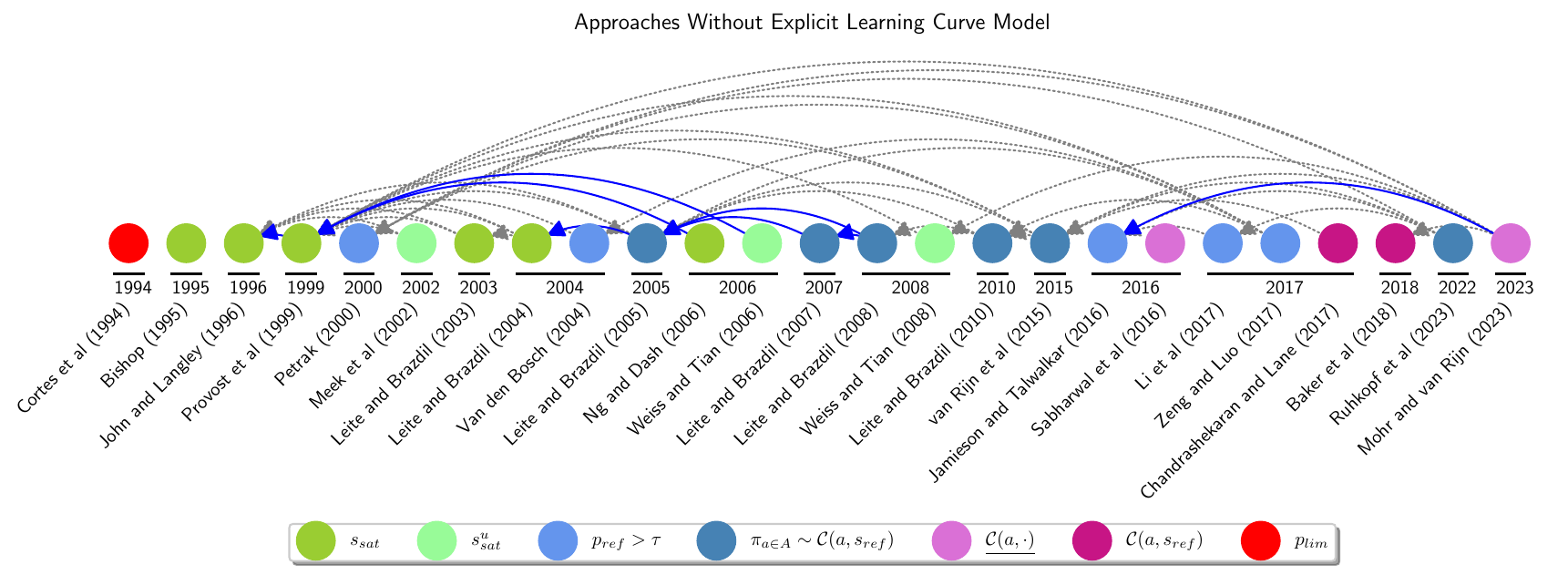}
    \caption{Citations across approaches without learning curve models. The colours indicate the question addressed about learning curves.
    Normal arrows indicate that the paper cited the other paper, and coloured arrows indicate an experimental comparison with previous approaches.}
    \label{fig:citations:withoutlcmodel}
\end{figure}

Many interesting questions related to learning curves can be addressed without even building an explicit learning curve model.
None of the questions in the lower layers in Fig.~\ref{fig:questions:learningcurve} necessarily requires a learning curve model.
Fig.~\ref{fig:citations:withoutlcmodel} shows a summary of all the approaches we are aware of, which make significant assertions or decisions related to learning curves without building a learning curve model.

\subsubsection{Prediction of Candidate Competitiveness (Binary)}
In this section, we discuss approaches that answer the early stopping criterion posed in Eq.~\ref{eq:early_discarding} without using a learning curve model.
The early discarding criterion can be seen as an instantiation of the binary question $\refperformance \leq \tau$, where $\tau = \min_{\learner^* \in A} \learningcurve(\learner^*, \refpoint)$ is the best value that any learner on the available resources.
This (conceptually simple) question is often answered by applying sophisticated learning curve models.
The usage of learning curve models is understandable since, at the time of the decision, neither \refperformance nor $\min_{\learner^* \in A} \learningcurve(\learner^*, \refpoint)$ is known; therefore, extrapolating learning curves offers a possibility to make assessments about these quantities.
However, it is also possible to say something about the early discarding condition without a learning curve model.

As far as we know, the only type of approach in this category is \emph{horizontal} early discarding with an \emph{implicit} affirmation of the early discarding criterion that only uses the last anchor of the empirical learning curve.
That is, during the model selection process, all remaining members of a candidate set \learnerspace are trained to some budget.
This budget can be either a sample size (\samplecurves), learning iterations, or time (\itcurves).
Then, some portion of candidates $\learnerspace^-$ is removed from \learnerspace based on the \emph{last value} of their respective curve.
The implicit assumption is that $\learningcurve(\learner, \refpoint) > \min_{\learner^* \in A} \learningcurve(\learner^*, \refpoint)$ for every $\learner \in A^-$.

To our knowledge, the first algorithm in this line was introduced in the \emph{Wrapped Progressive Sampling} procedure (WPS) by \citet{vandenbosch2004wrappedprogressivesampling}.
In this approach, a dynamically computed subset of the candidate set \learnerspace is discarded (instead of a constant fraction as 50\%).
WPS creates a histogram with ten bins $b_1,..,b_{10}$ of the candidates in \learnerspace based on their validation accuracy to decide which candidates are discarded.
Then WPS identifies the largest index $i^- = \max \{i \in \mathbb{N}, <10: \vert b_i\vert < \vert b_{i+1}\vert\}$ of a bin that has fewer elements than its successor.
Then, all candidates in bins with an index lower or equal $i^-$ are dropped.
In an extreme case, all except one candidate might be dropped right in the first iteration, e.g., if $\vert b_{10}\vert = 1$ and $\vert b_9\vert = 0$.

More recently, successive halving~\citep{jamieson2016nonstochasticbestarm} and hyperband~\citep{li2017hyperband} have been introduced, which are both adequate methods based on a simple concept.
Successive halving considers a set of candidate models \learnerspace, which all receive an initial budget. It iteratively drops 50\% of the candidate set \learnerspace while doubling the remaining candidates' budget until a single candidate model remains.
Hyperband is a series of successive halving \emph{brackets}, where each bracket is initialised with an increased initial budget and a new initial set of candidates.
\citet{zeng2017progressivesampling} proposed an extended version in which all the candidates that perform worse than the best candidate by some constant and pre-defined margin are discarded.
Interestingly, the above approaches operate without a learning curve model for extrapolation and refrain from using the existing observations other than comparing the last seen values.
Therefore, neither predictions nor observations of recent trends are being utilised. 
Nonetheless, these methods perform well empirically and come with theoretical guarantees. 

\subsubsection{Candidate Ranking}
\label{sec:literature:ranking}
The methods in this section aim to rank the learning algorithms with respect to their (expected) performance at the full dataset size.
In our scheme in Sec.~\ref{sec:taxonomy:technicalquestions} and Fig.~\ref{fig:questions:learningcurve}, we denote this question as \candidateranking.
Depending on the available resources (cf. Sec.~\ref{sec:taxonomy:resources}), such a ranking can be based on the other learning curves available from other contexts and the explicit characterisation of such contexts, e.g., features that describe datasets or algorithms.

\paragraph{Ranking Without Context Description}
The first approach we are aware of to address this problem was \emph{metalearning on data samples} (MDS)
presented by \citet{leite2005predictingrelativeperformance} through the notion of comparing two learners, i.e., a ranking of two.
Given a dataset, MDS decides which of two algorithms is the better choice on a given dataset. Therefore, it can be used for early discarding.
The authors specifically use an SVM and a C5 decision tree but rightfully claim that any algorithm could be used.
The formal basis of the work is the same as the one introduced in their previous work~\citep{leite2003improvingprogressivesampling,leite2004improvingprogressivesampling}.
Similar to previous work, it assumes that empirical learning curves (with standardised anchor sizes) for the learner under examination are already known for \emph{other} datasets. Additionally, it builds upon the idea of quantifying the \emph{distance} between the target and the other datasets based on the sum of squared distances over the already-known performances at anchors of the target dataset.
Once the most similar $k$ learning curves have been identified, MDS assigns a score to each learner that is the \emph{mean accuracy} of its $k$ nearest neighbours (at the final anchor).
It then selects the algorithm with the higher score.

\citet{leite2005predictingrelativeperformance} acknowledge that this method may result in poor rankings because even the closest learning curve on other datasets can still be substantially different and propose \emph{learning curve adaption} as a remedy.
Instead of forming the mean directly over the target anchors of the nearest neighbour empirical learning curves, the authors first \emph{scale} those curves to make them more similar to the shape already observed on the target dataset.
To this end, they compute a scaling constant under which the overall anchor-wise distance is minimised and then multiply all the scores with this constant.
This version of the MDS algorithm is called AMDS (probably for Adaptive MDS). 

Interestingly, one can argue that this adaption technique could be applied either \emph{before} or \emph{after} determining the $k$ nearest neighbours.
Doing it before could lead to other (and better) nearest neighbours because then the neighbours are determined more with respect to the \emph{shape} of the learning curve, and the \emph{offset} plays much less of a role.
However, in the above paper, the adaption is done \emph{after} retrieving the neighbours.

The authors extended the approach by creating an online sampling scheme with the SetGen algorithm~\citep{leite2007aniterativeprocess}.
SetGen is an online adaption of AMDS in that, after each acquired anchor, it is decided whether and which anchor should be evaluated with each algorithm; this considers both the (believed) accuracy and the runtime.
This procedure can be seen as a way of racing between the algorithms.
The potential of each additional amount of budget is judged based on the metalearning database.

An implicit assumption of all approaches in this line of research is that the datasets in the database from which performances are extracted need to be at least as big as the target dataset.
This issue was first explicitly treated in the \emph{Pairwise Curve Comparison} approach (PCC)~\citep{vanrijn2015fastalgorithmselection}.
This algorithm builds upon the works of \citet{leite2010activetesting} and implements a voting scheme to identify the best learning algorithm of a portfolio; votes are distributed based on wins, which are determined based on the predicted performance at the complete dataset \dataset.
\citet{vanrijn2015fastalgorithmselection} explicitly discuss the issue if the sizes of datasets in the database and the target dataset are not identical.
In general, the point for which predictions must be made is typically not one of the anchors; it is typically not a power of 2 but rather 90\% of the given dataset size (due to the holdout scheme).
A remedy is to resort to the closest available anchor in the schedule.
However, if the highest anchor available for another dataset is much smaller than the required training size of the target dataset, then it is unclear how that curve should be used.

\paragraph{Ranking With Context Description}
The first work we are aware of that realises an explicit context description was proposed by \citet{leite2008selecting,leite2010activetesting}.
Similar to the methods discussed earlier, these methods select the $k$ nearest datasets to measure the relevance of known complete learning curves of other datasets for performance prediction on the target dataset.
The main difference is that, in addition to the contribution of the partial learning curve itself to the distance, they also use the distance between the datasets in terms of their \emph{meta-features}.
More precisely, they compute the Manhattan distance between seven range-normalised dataset meta-features, e.g., dataset size, number of symbolic and numerical attributes, etc.
The overall distance between the datasets is then the \emph{sum} of the distance between the partial learning curves and the distance in terms of meta-features.
This work was marginally refined in the \emph{Selection of Algorithms using Metalearning} approach (SAM), which applies the same logic but assigns a \emph{weight} to each of the two distance sources~\citep{leite2010activetesting}; the weight is however implicitly assumed to be set to 0.5.

A recent and entirely different approach to candidate ranking is the MASIF transformer framework \citep{ruhkopf2022masif}.
This approach takes partial learning curves of different learners on the current task, which may have potentially different lengths and combines them with dataset meta-features in order to predict \emph{latent utility values} of each learner as expected for the complete dataset.
The transformer is trained based on previous experiences on datasets for which true rankings among the learners have been computed for the complete dataset.
It is unclear to which degree the utility values predicted by the transformer resemble the actual performance of the learners at the target size, but the paper suggests that the \emph{ordering}  \candidateranking of the learners according to these scores is relatively faithful to the true ordering induced by the actual performances.

\subsubsection{Identification of Saturation Performance}
Estimating the limit or saturation performance \satperformance is helpful in two decision-making situations:
\begin{enumerate}
    \item for \samplecurves, it can be used for data acquisition by checking whether the availability of more data promises to improve predictive performance significantly.
    \item for \itcurves, it can be used for early discarding. Given a threshold $p$, we can determine whether training a model to convergence will achieve a generalisation performance of at least $p$. 
\end{enumerate}
To the best of our knowledge, the question on estimates of \satperformance is the only of the above questions that received a substantial amount of theoretical contributions.
This is not surprising since \satperformance is an asymptotic quantity that is arguably suited for theoretical analysis.
The root of this line of research is the statistical mechanics framework~\citep{seung1992statistical}.
This and related research~\citep{murata1992learningcurves,amari1993statistical,fine99parameterconvergence} consider a type of capacity curve in which asymptotic properties of the learning curve are expressed in terms of the number of parameters, usually those of a neural network.
However, a side observation of these works is that there is a kind of symmetrical behaviour between the \emph{train error} and the \emph{validation error} (sometimes called generalisation error).

\citet{cortes1994limits} take these observations to use the mean of the two empirical curves to estimate \satperformance as soon as the training error starts to rise, i.e., as soon as the model cannot accommodate the training data perfectly anymore.
Although this work considers capacity curves to identify the intrinsic noise level of the data, i.e., the minimum error necessarily made by \emph{any} learner, they also report the asymptotic performance of a neural network on a single data set.

To the best of our knowledge, this is the only approach that estimates \satperformance without building an explicit learning curve model.
While several other methods are capable of estimating \satperformance of \itcurves of neural networks~\citep{swersky2014freezethawbo,domhan2015speedingup}, these rely on full learning curve models, which are therefore discussed in Sec.~\ref{sec:literature:performanceatanypoint}.

\subsubsection{Identification of Saturation Point}
Identifying the saturation point \satpoint of \samplecurves is useful in the following cases:
    \begin{enumerate}
    
        \item Early-Stopping with \samplecurves: Which \emph{portion} of the \emph{available} data is necessary to obtain saturation performance?
        
        This is relevant if $\vert\dataset\vert > \satpoint$ or the relationship between $\vert\dataset\vert$ and \satpoint is unknown and training on full \dataset is potentially undesirable.
        
        \item Early-Stopping with \itcurves: How many iterations are necessary until performance converges?
        This applies to incremental learners, such as neural networks.
        
        \item Data acquisition:
        How many additional labelled observations are necessary to obtain (near-optimal) performance?
        This applies if $\vert\dataset\vert < \satpoint$.
        The question can be posed for a specific learner or a portfolio.
    \end{enumerate}
    
\paragraph{Retrospective Approaches}
The simplest way of determining the saturation point \satpoint is to incrementally build a learning curve and stop as soon as it is believed that the saturation point has been \emph{exceeded}.
If we do this, we can estimate that the saturation point lies between the last two anchors.
For \itcurves, determining \satpoint comes for free as a side-product of the training procedure. 
It is commonly used for training neural networks~\citep{bishop1995regularization,goodfellow2016deep}.
On the other hand, it requires restarting and is potentially costly for \samplecurves.

\citet{john1996staticvsdynamic} define a \emph{dynamic sampling} approach to determine the \satpoint for \samplecurves.
A straightforward approach mentioned in that paper is to observe whether the performance has become \emph{worse} on the last sampled anchor.
If so, one might consider gathering empirical evidence that the saturation point has been \emph{exceeded}, i.e., it should be somewhere between the last two anchors.
However, the authors argue that preliminary results indicate that this approach often stops too early.
This is mainly caused by high aleatoric uncertainty, which implies noisy empirical learning curves.
On the other hand, one could argue that this approach stops far too \emph{late} because it can require quite some iterations until, by chance, the observed performance is worse than the one of the last iteration.
Therefore, \citet{john1996staticvsdynamic} propose also a model-based approach to avoid this problem, which we discuss in Sec.~\ref{sec:literature:performanceatanypoint}.
Of course, when having access to such a model, we can query the expected performance and compare it to the performance at the last anchor.

\citet{provost1999efficientprogressivesampling} address the stability issues and also (some of) the efficiency issues of the above trivial approach in a scheme they call \emph{progressive sampling}.
Similar to dynamic sampling~\citep{john1996staticvsdynamic}, progressive sampling induces models for each anchor in the schedule until the convergence of the learning curve is detected.
There are two main differences between the two approaches.
First, the authors propose to use \emph{geometrical} instead of arithmetic schedules, i.e., a schedule of the form $b^k$ instead of $bk$, where $b$ is a constant and $k$ is the position of an anchor in the schedule.
They prove that every geometric schedule is asymptotically optimal in terms of runtime; that is, every such schedule has the same asymptotic runtime as the schedule that evaluates only on \satpoint.
This optimality proof only holds in the asymptotic calculus; in practice, there are better and less good geometric schedules.
For example, \citet{provost1999efficientprogressivesampling} propose a dynamic programming approach (called DP), which efficiently computes the cost-optimal schedule based on a prior distribution on \satpoint and a given training runtime model.
The second difference is that \citet{provost1999efficientprogressivesampling} check whether the saturation point has been reached using a method called \emph{linear regression with local sampling} (LRLS), which samples not only \emph{at} but also closely \emph{around} anchors to estimate the slope at an anchor and stop if the slope is close to zero.

The practical benefit of the LRLS scheme is not entirely clear for \samplecurves.
First, since the slope is also based on empirical values, from a theoretical viewpoint, it is not clear that the criterion is necessarily better than the naive approach suggested by \citet{john1996staticvsdynamic}.
Second, the approach is more expensive than the naive approach because more observations need to be sampled; this can be a substantial factor, especially for large anchors.
LRLS becomes relevant when applied to learners that have a computational complexity for training that scales worse than linear in the number of training points (e.g., Gaussian processes or decision trees). 
For learners with training complexity linear in the number of observations, it will often be more expensive than evaluating a learner's performance on the complete dataset.
This can be seen with a simple calculation, in which we assume roughly linear training time complexity:
Suppose a costly anchor at 40\% of the overall data size.
If we draw only two additional samples around this anchor, then the runtime is around 120\% of the runtime we would have had if we had trained on the complete dataset once.
While \citet{provost1999efficientprogressivesampling} argue that LRLS only adds a constant factor to the runtime, \citet{sarkar2015costefficientsampling} reasonably argue that this factor is often prohibitive in practice.

The approaches in this section explicitly assume that more data than \satpoint is available.
This implies that they can be used primarily for early stopping scenarios rather than data acquisition scenarios. 
Still, if a learner does not attain saturation performance on the complete dataset, the approaches can detect this at the cost of the additional evaluations at the non-final anchors.

Concerning the stability of estimates, \citet{beleites2013sample} point out the necessity to have an estimate for the \emph{confidence interval} not only for the performance at the anchors in the training schedule but also on the \emph{test data}.
They argue that confidence intervals are essential when deciding whether reasonable generalisation statements can be made for a classifier.
This changes the notion of the stopping point to, perhaps, a \emph{confident} stopping point.
The optimal stopping point may be reached early, but the validation fold sizes may still be too small to assure stable assertions.
Typically, the confidence intervals are large on small anchors and then contract for increasing anchor sizes.
Based on credible intervals, the authors propose choosing the anchor point that achieves a sufficiently narrow interval on the test data.

\citet{ng2006progressiveoversampling} address the impact of class imbalance on the performance of a learner.
That approach hypothesises that, without further knowledge, the class distribution in which all classes have the same number of observations is optimal.
The authors modified the aforementioned progressive sampling scheme by creating train sets at each anchor such that all classes have the same distribution.
For anchors of sizes that would require more instances of a class than are available in the existing data, random instances of that class are replicated until the class balance is established again.
Note that while the work is based on the findings by  \citet{weiss2003learningwhentrainingdatarecostly} (cf. Sec.~\ref{sec:literature:classdistribution}), they apply a different strategy.
Instead of optimising over the class distribution for a given anchor size, they try to find the stopping point under the premise that the training set will always be balanced.

Regarding the stopping point of \itcurves, a common technique is to separate some data that is not used for training but to compute an \itcurve online to detect convergence~\citep{bishop1995regularization}.

\paragraph{Projective Approaches}
A different idea to obtain the saturation point \satpoint was proposed by \citet{leite2003improvingprogressivesampling,leite2004improvingprogressivesampling} through the notion of metalearning~\citep{brazdil2022metalearning}.
Similar to \citet{provost1999efficientprogressivesampling}, a geometric schedule is used.
The assumption is that we already know the performances of the current learner at all anchors in the schedule on \emph{other} datasets.
The idea is to compute, on the target dataset, the performances only for the very first anchors and then to \emph{predict} the saturation point by aggregating the (known) saturation point on the $k$ \emph{most similar} learning curves of the other datasets.
The distance measure here is the sum of differences between the curves at the initial anchors; the concrete anchors used in their paper are (91, 128, 181, 256, \ldots), corresponding to the powers of $\sqrt{2}$.
The authors consider different aggregation measures such as mean and minimum~\citep{leite2003improvingprogressivesampling} and the median~\citep{leite2004improvingprogressivesampling}.

The authors discuss the potential issue that, among the $k$ nearest neighbour curves, some or even all of the curves can be substantially different from the partial learning curve on the target dataset.
Using the $k$ nearest learning curves to predict the stopping point would not work in such cases.
Follow-up work~\citep{leite2007aniterativeprocess} proposes a remedy to this problem, in which the curves are not used directly but are adjusted via a concept called \emph{curve adaptation}
(discussed in Sec.~\ref{sec:literature:predictionatfixedpoints:contextfreegeneralization}).

\subsubsection{Finding the Utility-Based Stopping Point}
\label{sec:literature:economicstoppingpoint}
The problem of identifying the utility-based stopping point was, to our knowledge, first addressed by \citet{meek2002thelcsamplingmethod} and was also independently investigated by  \citet{weiss2006maximizingclassifierutility}.
In these papers, a retrospective approach is applied.
The idea is similar to the aforementioned concept of progressive sampling~\citep{provost1999efficientprogressivesampling}, except that the analysis is done for utility rather than learning curves.
In contrast to the learning curve, the utility curve does not plateau but starts to deteriorate after its peak (see Fig.~\ref{fig:performancecurves_utility}).

The main difficulty with the concept of utility in the context of learning curves is to find a unifying scale for (i)~the costs of data acquisition and training time and (ii)~the model performance.
\citet{meek2002thelcsamplingmethod} avoid this problem by adopting the notion of implicit utility through the comparison with a baseline.
They stop the algorithm when the ratio between the benefit improvement and the augmented runtime drops below a pre-defined threshold.

In contrast, \citet{weiss2006maximizingclassifierutility} compute an explicit utility, and the algorithm stops as soon as the \emph{observed} utility decreases for the first time, which is taken as the indicator that the utility-based stopping point has been passed.
The authors adopt the concept of the net utility of a potential classifier, which is the difference (in utility) between predictive performance (under a hypothetical number of training instances) and the cost to acquire the (additional) instances.
Notably, the assumption is that the user has no control over the instances for which labels will be acquired next, which contrasts the approach from \emph{active learning}.
To merge different types of inconveniences (i.e., acquisition costs and prediction errors) into a single utility measure, the user has to \emph{define} costs per unit, e.g., costs for acquiring a single usable training instance and costs for making a wrong prediction.
Furthermore, the authors consider the problem of deciding \emph{online} whether or not to acquire more data and, in the affirmative case, how \emph{many} instances should be considered in the acquisition batch before reconsidering.
The latter effectively corresponds to deciding upon a progressive sampling scheme~\citep{provost1999efficientprogressivesampling}.
However, the paper does not analyse the effects of fixed costs per batch (such as the computational costs of training a model), which implies that one could set batch sizes to 1 without consequence.

A consecutive version of that paper also adds the CPU cost for model induction to the costs of a point on the learning curve ~\citep{weiss08maximizingclassifierutility}.
The original paper only considered acquisition costs and the prediction error.
This model also seems suitable for \itcurves, where there would be no data acquisition costs.

In all of the above approaches, the usage of the empirically gathered learning curves for decision making is minimal.
Moreover, the approaches ignore all except the \emph{last two} points on the learning curve.
In this sense, and in terms of the stopping point approaches, the above works are retrospective in nature.
No model of the learning curve is built, and no projections of errors on bigger training sizes are made, which, for example, could make sense to predict that this utility peak event will occur in the future or even only in the next iteration.
\citet{last2007predicting} proposes such an approach, which we discuss in Sec.~\ref{sec:literature:utilityprediction}.

\subsubsection{Performance Prediction at Fixed Point}
\label{sec:perfpredict}
The problem of predicting the learning curve value is naturally a regression problem, where the goal is to predict \curvemeanab for a fixed budget \budget.
Essentially, given a fixed learner \learner and the budget expressed in either observations \sampleanchor or iterations \timeanchor, it is about predicting $\learningcurve(\learner, \sampleanchor)$ for \samplecurves, and predicting $\learningcurve(\learner,\sampleanchor,\timeanchor)$ for \samplecurves.
The attributes are the performance values at different (cheap) anchors and potentially additional contextualising attributes. 
Using these attributes, one explicitly or implicitly generalises over \emph{datasets} or \emph{learning algorithms} (or both).
Models that generalise over datasets are typically called meta-models and rather aim at model ranking~\citep{brazdil2022metalearning,ruhkopf2022masif}, which we discuss in Sec.~\ref{sec:literature:ranking}.
This is because it is difficult to generalise exact performance across datasets.
When generalising over learners, one typically trains a single model for the target dataset, trained from learning curves on the same dataset belonging to other learning algorithms. 
Models that generalise over learners are typically called surrogate models~\citep{eggensperger2018efficient}.

We organise this section by how explicit the generalisation is made over one of the two concepts.
If no additional attributes are available, one implicitly assumes that all entries in the database are to a degree suitable to predict values in a new situation.
There is no explicit context in this case, and we are generalising implicitly over datasets or learners.
Since no explicit contextualisation exists, those approaches can be used for both purposes, regardless of their original purpose.
In contrast, additional contextualising attributes can describe the dataset, i.e., we have meta-features of the datasets available, or they can describe the learning algorithms to which the learning curve values belong.
In principle, one could utilise both types of additional attributes, but we are not aware of any approaches that adopt both.

\paragraph{Generalization Without Explicit Context}
\label{sec:literature:predictionatfixedpoints:contextfreegeneralization}
Generalisation from learning curves without explicit context means to predict the performance of a given learner on some dataset based on previously acquired learning curves that are not equipped with additional information, i.e., features describing the dataset or the algorithm used to produce them.
To justify the prediction model, the existing empirical learning curves either stem from the same algorithm on other datasets, or other algorithms on the same dataset. 
Either of these implicitly qualifies them to be relevant for the new task.
In other words, one simply uses a set of unannotated existing learning curves as aids to predict the behaviour of a new, only partially known, learning curve.

\citet{chandrashekaran2017speedinguphpo} developed an approach that explicitly used regression to predict the target performance without context.
Probably without noticing, the approach mainly re-invents the approaches previously developed by \citet{leite2005predictingrelativeperformance,leite2007aniterativeprocess,leite2010activetesting,vanrijn2015fastalgorithmselection}, since it computes the most similar other learning curves in the portfolio and obtains a prediction based on the average over those curves.
The three differences are that \citet{chandrashekaran2017speedinguphpo} (i)~consider uncertainty in the prediction based on the variance in the neighbourhood, (ii)~adopt an affine transformation (instead of a linear transformation) of the existing learning curves and apply this \emph{before} selecting the most similar ones, and (iii)~that they do not use a fixed schedule but, due to the focus on \itcurves, simply a continuous schedule that is stopped as soon as there is enough evidence that the target performance will not be better than a current threshold.
The Euclidean norm between the vectors describing the performances at the anchors is used as the distance function between two curves.

The reason why the approach is discussed here and not in Sec.~\ref{sec:literature:ranking} together with the others is subtle and worth being discussed.
In both lines of research, target performance values from related curves are averaged to estimate the performance of the current learner.
However, \citet{chandrashekaran2017speedinguphpo} explicitly treat this as a performance prediction, which is then used for early discarding by comparing against a threshold.
In contrast, the works in the line of \citet{leite2005predictingrelativeperformance} do not treat this value as an actual prediction but simply as some score used to order a pair or a set of learners.

While the approach generalises across algorithm configurations, it ignores the configurations and generalises from learning curves without explicit context.
In other words, there is no reason why the approach could not be used also for generalisation across datasets.

\citet{cardonaescobar2017efficienthpo} presented an approach that predicts values at \emph{all} anchors.
The authors adopt a series of support vector regression models, one for each anchor not evaluated so far.
It is not entirely clear with which data the models are trained, but we presume that it follows the same logic as \citet{chandrashekaran2017speedinguphpo} and uses the fully known learning curves of previously evaluated neural network configurations to do so.
Interestingly, similar to chaining \citep{gkioxari2016chained} in classification, they use as inputs for the $j$-th future anchor not only the known partial learning curve values but also use the \emph{predictions} for the anchor points predicted before~$j$.

\paragraph{Generalization With an Explicit Algorithm Context}
Generalisation across algorithms only considers the target dataset and assumes that a number of (complete) empirical learning curves on that dataset are already available for different algorithms.
The \emph{explicit} generalisation requires that the previous learning curves are \emph{explicitly} associated with features describing the algorithm to which they belong.

\citet{baker2018acceleratingnas} propose a method that uses features describing \emph{both} the learning curve (including up to second-order differences) \emph{and} the algorithm.
The approach predicts the performance of neural networks based on features that describe the architecture (number of layers and weights) as well as the hyperparameters of the learning algorithm (such as learning rate, learning rate decay, etc.).
They adopt linear and kernel-based support vector regression machines, random forests, and simple linear regression based on ordinary least squares.
Even though the authors suggest using kernel-based support vector regression machines, they find that simple linear regression also often compares highly competitive for this prediction task.

Following this idea, \citet{long2020performance} additionally add textual descriptions of the architecture to predict the learning curves of neural networks.
Indeed, the architecture description by \citet{baker2018acceleratingnas} is rather simplistic and only immediately well suited if all layers are of the same type, e.g., dense layers, and have the same number of neurons.
\citet{long2020performance} report substantial performance improvements for convolutional network architectures compared to the approach taken by \citet{baker2018acceleratingnas}.

\subsubsection{Performance Bounding}
\label{sec:literature:anypoint:bounds}
Performance bounding tries to give explicit lower or upper bounds on the performance value at some specific or arbitrary budget.
Answering this question is essential to make high-confidence decisions on early discarding as discussed in Sec.~\ref{sec:decisionsituations:earlydiscarding}; formally, we denote it as $\underline{\learningcurve(\learner, \cdot)}$.
Typically, one is interested in $\underline{\learningcurve(\learner, \refpoint)}$, $\refpoint$ being the size of the dataset intended for training or the limit performance of an \itcurve, but the models we discuss here are more general.

Performance bounding is intuitively a simpler problem than performance prediction (Sec.~\ref{sec:perfpredict}), but there are usually also higher expectations with respect to the accuracy of the assertion.
When a performance bound is expressed, one would expect that the true value is at least as high as specified with high probability.

One approach that addresses this problem is Data Allocation using Upper Bounds (DAUB) ~\citep{sabharwal2016selecting}. 
Given a set of configurations, it first runs all configurations on two anchors of the dataset, effectively building the initial segment of the learning curve.
Based on this initial segment per configuration, it determines an optimistic performance bound for each learning curve that likely will not exceeded (i.e., an upper bound for measures that need to be maximised, such as accuracy, and a lower bound for measures that need to be minimised, such as error rate).
This performance bound is determined by calculating the linear regression slope of the last two segments. The performance on the last anchor is extrapolated to the full size of the dataset according to this slope.
Therefore, there is an optimistic upper bound on the performance that a configuration can obtain.
After that, it goes into the following loop:
It runs the most promising configuration on a larger sample size and updates the performance bound.
It reevaluates which configuration has the most potential at that budget and assigns more budget to the most promising configuration until one configuration has been run on the entire dataset. 
Therefore, this is an example of horizontal model selection.  

Alternatively, learning curve cross-validation uses a similar approach but addresses this in a more flexible, vertical setting~\citep{mohr2023fast,mohr2021towards}.
The method explicitly assumes that \samplecurves have a convex behaviour.
The convexity of the curve allows for deriving a best-case extrapolation from a partial empirical learning curve.
The convexity assumption is used to linearly extrapolate the empirical learning curve and prune learners when they can no longer improve on the best-known solution.

\subsection{Approaches With Learning Curve Models}
\label{sec:literature:withmodel}
\begin{figure}[t]
    \centering
    \includegraphics[width=\textwidth]{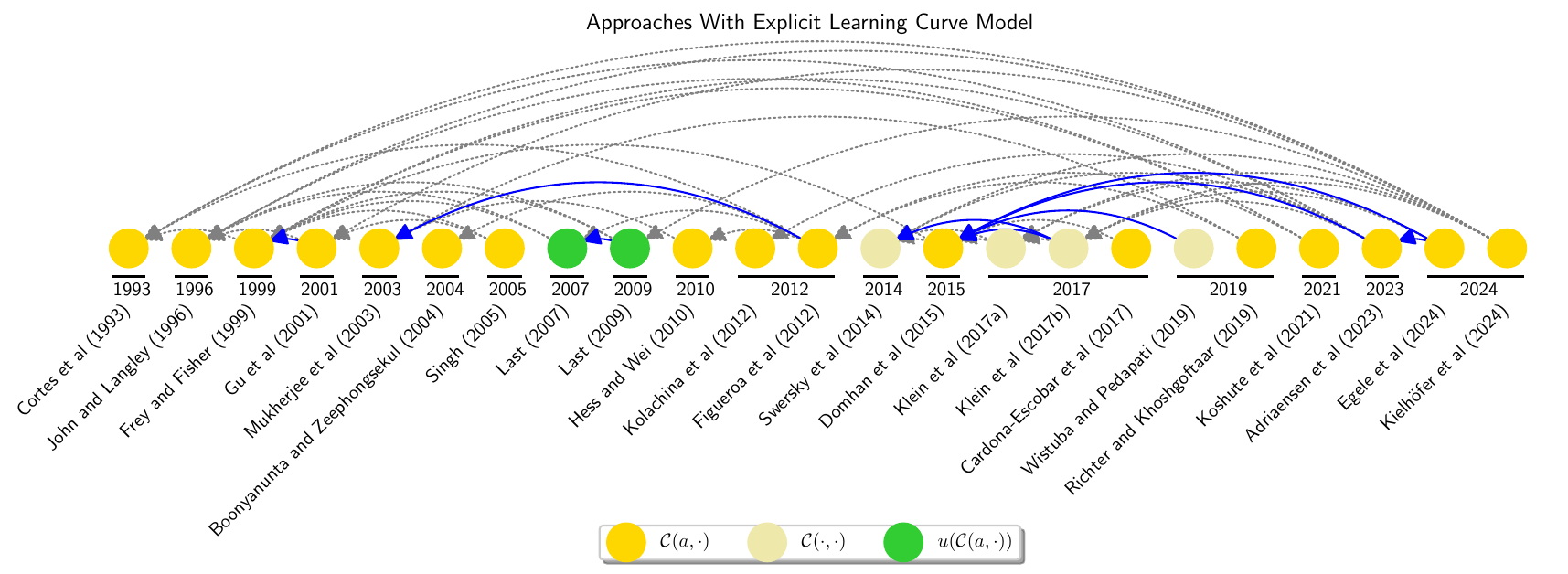}
    \caption{Citations across approaches with a learning curve model. The colours indicate the question addressed about learning curves.
    Normal arrows indicate that the paper cited the other paper, and coloured arrows indicate an experimental comparison with previous approaches.}
    \label{fig:citations:withlcmodel}
\end{figure}

This section covers learning curve approaches that utilise an explicit model to model the entire learning curve, as described in Sec.~\ref{sec:curvemodels}.
Fig.~\ref{fig:citations:withlcmodel} shows an overview of all the approaches we discuss and how they cite each other.
In the base form, there is a model for a specific learner \learner that is able to predict the performance $\learningcurve(\learner,\cdot)$ or $\learningcurve(\learner,\sampleanchor,\cdot)$ at any sample size or iteration respectively (yellow), discussed in Sec.~\ref{sec:literature:performanceatanypoint}.
On top of such curves, a utility curve \utilitycurve can be defined (green), discussed in Sec.~\ref{sec:literature:utilityprediction}.
Finally, the curve models can even generalise across the learners, leading to generalised curve models $\learningcurve(\cdot, \cdot)$ (beige). 
One additional benefit of these types of models is that they can also perform model acquisition, i.e., assess the performance of a learner for which we have no performance evaluations yet. These models are discussed in Sec.~\ref{sec:literature:anypointandanylearner}.

In this section, we describe the learning curve model according to the notion of \emph{budget}, using the variable \budget to avoid having to distinguish between sample sizes \sampleanchor or iterations \timeanchor.
In the spirit of Sec.~\ref{sec:curvemodels}, we use the notation \lcab to refer to the original random variable that generates observations of the performance of learner \learner at budget \budget (independently of whether \lc here is a \samplecurve with \budget being the sample size or whether \lc is an \itcurve with some implicit sample size \sampleanchor and \budget being the iteration).
Accordingly, $\curvemeanab = \mathbb{E}[\lcab]$ is the mean value of the curve of learner \learner at budget \budget.

\subsubsection{Performance Prediction at Any Point}
\label{sec:literature:performanceatanypoint}
We will discuss approaches that build a full learning curve model \lcestb for a specific learner \learner.
That is, they model the curve mean \curvemeanab for a fixed learner \learner at \emph{any} budget \budget (effectively modelling \curvemeana).
We divide the approaches into three groups, corresponding to the three model types for \lcest discussed in Sec.~\ref{sec:curvemodels}, which are also reflected in the three layers of Fig.~\ref{fig:questions:learningcurve}.
Accordingly, we first discuss approaches that provide \emph{point estimates} of the curve, i.e., $\lcest_\learner: \mathbb{N} \rightarrow \mathbb{R}$, then approaches that explicitly treat those estimates with uncertainty and introduce a notion of bounds on them, i.e., $\lcest_\learner: \mathbb{N} \rightarrow \mathbb{R}^2$, and finally approaches that create entire probabilistic belief models over curves, i.e., $\lcest_\learner: \mathbb{N} \rightarrow \{p~\vert~ p \text{ is a distribution of a real-valued random variable}\}$.

\paragraph{Point Estimates}
\label{sec:literature:anypoint:point}
To our knowledge, the first approach that used observed data to fit a learning curve model was presented by \citet{cortes1993learningcurves}.
In that paper, the three-parametric inverse power law shown in Eq.~(\ref{eq:ipl}) was used to build an \itcurve.
The usage of the power law model is justified with the findings in the statistical mechanic framework~\citep{seung1992statistical} and used to predict the predictive performance on the complete dataset (for two neural network architectures on the NIST dataset). 
The authors find that the predictive performance on 60k instances can be almost perfectly predicted using the inverse power law.
Unfortunately, it is not entirely clear on which anchors the estimates are built.
Notably, this is the only paper we are aware of that fits \emph{both} a train- and test-error curve.
The paper reports some information about noise through boxplots on the curve but does not explicitly incorporate them into the model.

\citet{john1996staticvsdynamic} proposed a similar approach.
Similar to the work of \citet{cortes1993learningcurves}, the model is used to estimate the performance of a learner on the complete dataset with the goal of stopping the training procedure early.
One difference is that the model is adopted for a \samplecurve instead of an \itcurve.
Concerning convergence detection, \citet{john1996staticvsdynamic} employ the Probably Close Enough criterion, which detects convergence if the probability that the accuracy of a model trained on the complete dataset will be at most some $\varepsilon$ worse than the current model's performance is less than some $\delta$.
However, the paper itself then does not adopt a notion of probabilities but stops if the performance on the complete dataset predicted by an inverse power law model is not by at least $\varepsilon$ better than the currently observed performance; they call the approach Extrapolation of Learning Curves.
In other words, the uncertainty is not quantified.

The inverse power law has been used in many applications.
For example, several approaches have used the inverse power law to model the performance of neural networks in different domains~\citep{cho2015much,alwosheel2018isyour}.
It is noteworthy that recent works show evidence against the usage of any commonly used models for neural networks, at least in the initial parts of the curve, due to the (sample-wise) double descent~\citep{nakkiran2020deepdoubledecent}.
At least for certain combinations of architectures, datasets, and training procedures, there is empirical evidence that the learning curve exhibits non-monotonic behaviour, which contradicts all existing learning curve models like the inverse power law model.

\paragraph{Range Estimates}
\label{sec:literature:anypoint:ranges}
\citet{mukherjee2003estimatingdssizerequirements} built inverse power law models in the domain of DNA data.
The main contribution of that paper is to analyse the appropriateness of the inverse power model on eight medical datasets.
To this end, they construct uncertainty bounds around the mean learning curve consisting of the q25 and q75 curves fitted from those statistics, respectively.
This way, a learning curve model including information about dispersion is obtained.
Experiments are conducted for a support vector machine on eight medical datasets in which the leave-one-out validation result (estimate of the learning curve on $\vert\dataset\vert - 1$ data points for training) is compared to the boundaries suggested by the model.
Having an explicit model for the q25 and q75 curves, one can obtain for an arbitrary anchor \budget not only a point estimate of \curvemeanab but an estimate of the inter-quartile range of \lcab itself, which is arguably more informative given that, assuming Gaussian noise, the interval should also contain \curvemeanab.

\citet{figueroa2012predictingsamplesize} modify the aforementioned approach in two ways.
First, different anchors are associated with different \emph{weights}, usually to assign higher weights to larger anchors since they are more informative.
Second, implicitly assuming a standard Gaussian distribution of the observations as in Eq.~(\ref{eq:noisemodel}), they compute a 95\% confidence interval to describe the uncertainty rather than the interquartile range.
Based on this information, for a query point \budget, they predict a confidence interval instead of a point estimate.
\citet{figueroa2012predictingsamplesize} also applied this approach to medical data, just as \citet{mao2016predicting} for EEG data.
More recently, it was also successfully used for sensor communication~\citep{oyedare2019estimating}.
This work aims to predict a reasonable sample size, which is perhaps more reasonably addressed by the utility-based approaches discussed in Sec.~\ref{sec:literature:economicstoppingpoint}.

Recently, \citet{koshute2021recommending} have used the inverse power law to predict the minimum anchor point on which a learner must be trained to reach near-\satperformance performance with a \emph{given desired} confidence.
This approach can be seen as a combination of the above two approaches.
Similar to \citet{figueroa2012predictingsamplesize}, they compute the confidence interval at all anchors.
However, instead of using these to estimate confidence intervals at arbitrary points, they fit a single curve on the lower bounds of the confidence intervals at the known anchors.
The resulting model is not used to make predictions on arbitrary anchors but to compute the cheapest anchor that will obtain with a pre-defined probability (size of confidence-interval) a performance that is $\varepsilon$-close to \satperformance, where $\varepsilon$ is a hyperparameter controlled by the user.

The idea of computing confidence intervals is also adopted by learning curve cross-validation~\citep{mohr2023fast,mohr2021towards}.
A Morgan-Mercer-Flodin model is created to decide whether or not to skip intermediate anchors and evaluate the learner on the full dataset size.
However, the confidence intervals are used differently than in the above cases and are not used for the inverse power law model itself.
In contrast, the confidence bounds are used to compute the range of possible \emph{slopes} of the learning curve between two anchors.

\paragraph{Distribution Estimates}
\label{sec:literature:anypoint:distributions}
The first approach to predict distribution estimates for any anchor point was presented by \citet{domhan2015speedingup}.
The approach assumes learning curves to be instances of a parametric model that is a \emph{linear combination} of known model classes, such as the inverse power law, and others~\citep{gu2001modelingclassificationperformance}.
The main difference to the above approaches is that, instead of estimating the parameters through a maximum likelihood approach, they estimate, for each parameter, the whole posterior \emph{distribution} adopting Monte-Carlo Markov Chains (MCMC).
The approach is successfully used to early discard neural network architectures by predicting the saturation performance \satperformance of an \itcurve, thereby discarding learners as soon as the probability that it is competitive drops below a pre-defined threshold.
The Bayesian model proposed by \citet{domhan2015speedingup} also explicitly estimates the noise \curvenoiseab, which is assumed to be homoscedastic, i.e., identical for all budgets \budget.
In this sense, the approach quantifies \emph{both} epistemic and aleatoric uncertainty (cf. Sec.~\ref{sec:curvemodels}).

A recent alternative for estimating distributions has been proposed by \citet{adriaensen2023efficient} through the notion of prior fitted neural networks (PFN)~\citep{hollman2023tabpfn}.
Based on the approach by \citet{domhan2015speedingup}, \citet{adriaensen2023efficient} consider a set of basis functions, over which they define a prior.
They then sample a large number of curves from this prior and train a variation of a transformer neural network with it, which is able to predict \emph{distributions} for a target anchor based on a partial learning curve.
Since no sampling from the posterior is required anymore, compared to the MCMC approach by \citet{domhan2015speedingup}, predictions can be obtained much faster, and the authors claim that the prediction performance is comparable or even better, which could however not be confirmed in subsequent experiments discussed below.

The latest approach we are aware of is the robust estimation RoBER presented by \citet{egele2024unreasonable}.
The approach follows the idea of MCMC introduced by \citet{domhan2015speedingup} but applies a different sampling algorithm to obtain more stable estimates.
The results suggest that classical learning extrapolation significantly outperforms PFN-based extrapolations at the current state of research.

\subsubsection{Utility Prediction at Any Point}
\label{sec:literature:utilityprediction}
Utility prediction combines learning curve models as discussed in Sec.~\ref{sec:literature:performanceatanypoint} with utility models as discussed in Sec.~\ref{sec:literature:economicstoppingpoint}.
The learning curve model \lcest is used as a basis to estimate the performance at any point, and the utility at budget \budget is then computed as a function of the modelled performance  $\lcest_\learner(\budget)$ at and the associated costs for budget \budget.

The first approach in this direction was presented by \citet{last2007predicting}.
This work is very similar to the works of \citet{weiss2006maximizingclassifierutility} but makes utility \emph{forecasts} rather than looking back.
Therefore, it is projective instead of retrospective.
The error rate that serves as input to the model is obtained from a parametric model (i.e., a power-law) trained on the empirical values obtained at earlier budgets.
This framework enables one to \emph{analytically} compute the optimal dataset size.
The main advantage of the approach over the one of \citet{weiss2006maximizingclassifierutility} is that one does not need to go through several acquisition iterations, which is a benefit if those are associated with fixed costs.
Therefore, the learning curve has become a resource for decision making.
While this work assumes the empirical learning curve to be available, \citet{last2009improving} embeds his idea into an algorithm that follows a progressive sampling scheme in a follow-up work.

One use case in which the above techniques have been adopted has been reported in the context of automated software configuration~\citep{sarkar2015costefficientsampling}.
The context of that paper is that every instance is a parametrisation of a software library, and obtaining its label requires the costly execution of a benchmark on such a configuration.
The goal is to understand how many observations must be acquired to learn a reliable prediction model.
To this end, the authors adopt the projective sampling approach of \citet{last2007predicting,last2009improving}.

The latter work raises an important issue by stating that knowing the saturation point is (often) not enough.
Instead, we often also \emph{need} to know the performance at the saturation point.
\citet{sarkar2015costefficientsampling} argue that if the user is unaware of the expected performance at that point, substantial resources might be required to obtain the observations to reach the saturation point.
However, if the actual performance at that point is known to be mediocre, the user could anticipate this and not invest the required resources.
To this end, they also incorporate the utility model proposed by \citet{weiss08maximizingclassifierutility}.

\subsubsection{Performance at Any Point for Any Learner}
\label{sec:literature:anypointandanylearner}
The approaches discussed in this section are the most general ones developed to date regarding learning curves in
that they create a model for the complete function \learningcurve, i.e., generalising both over both budgets and learners.
Such a model is so versatile that it can be used in \emph{all} types of decision-making situations, e.g., data acquisition, early stopping, and early discarding. Additionally, these can be used for model acquisition (selecting a yet unseen promising model). 

Freeze-thaw Bayesian optimisation models the behaviour of learning curves through Gaussian processes \citep{swersky2014freezethawbo}.
An important contribution of that work is a non-stationary kernel for Gaussian processes that supports exponentially decaying learning curve models; it can easily be checked that standard kernels like a linear or Gaussian kernel do not lead to meaningful learning curve models.
Assuming that the kernel reflects the model class appropriately, one additional benefit of using a Gaussian process is that one automatically obtains estimates for the noise \curvenoiseab at an arbitrary anchor \budget.
Using their kernel and the current set of observations, \citet{swersky2014freezethawbo} estimate the asymptotic mean performance \satperformance.
Since the learning curves are combined with Bayesian optimisation, the uncertainty for a specific future anchor is one of the required inputs for computing their acquisition function.
In a rather thin evaluation, the approach was successfully applied to Online Latent Dirichlet Allocation, Logistic Regression, and Probabilistic Matrix Factorization, considering one dataset per learner.
While the paper focused on \itcurves, the modelling technique can also be used for \samplecurves.

\citet{klein2017fabolas} presented a similar approach dubbed FABOLAS.
Similar to Freeze-thaw Bayesian optimisaion, a Gaussian process is used to model the learning curve of the learners across different hyperparameter configurations.
There are three main differences between the two approaches as far as learning curves are concerned.
First, FABOLAS considers \samplecurves, while freeze-thaw Bayesian optimisation considers \itcurves.
Second, and related to this, FABOLAS uses a kernel different from freeze-thaw Bayesian optimisation to model the behaviour of the learning curve using the Gaussian process, which is defined by the relative dataset size in $[0,1]$ instead of absolute sizes.
Third, FABOLAS tries to explicitly learn the complete learning curve, while freeze-thaw Bayesian optimisation focuses only on the saturation performance.
Similar to freeze-thaw Bayesian optimisation, the uncertainty about the performance estimates of the learning curve is not explicitly used.
Still, the fact that a Gaussian process is fitted from the data allows one to make assertions about the certainty of the learning curve value at any point.

Parallel to their work on FABOLAS, \citet{klein2017lcpredictionwithbnn} proposed an approach to estimate both the mean and the \emph{noise} (aleatoric uncertainty) of a learning curve through the notion of Bayesian neural networks.
The neural network predicts the parameters of a set of basis functions. 
These basis functions incorporate prior knowledge into the network, which is necessary to extrapolate away from the data.
The main difference between this approach and the aforementioned approaches is that this approach models the behaviour of the learning curve through a neural network.
This network has $d+1$ input units ($d$ for the algorithm description and one for the anchor), one output unit for the estimated performance and, optionally, one output unit for the estimate of the variance of the performance (which relates to aleatoric uncertainty).
To our knowledge, this approach and the learning curve extrapolation proposed by \citet{domhan2015speedingup} are the only approaches in this area that explicitly model the variance of the performance of the learning curve (i.e., which can also be seen as noise).
An important difference between the two is that \citet{klein2017lcpredictionwithbnn} assume \emph{heteroscedastic noise}, i.e., noise that changes with both different hyperparameters and anchors, while \citet{domhan2015speedingup} assume homoscedastic noise across anchor sizes (not across configurations, because the model does not generalise over different configurations).
While the approach presented in the paper does not explicitly consider the uncertainty about the parameter estimates, the parameters are essentially sampled from a posterior distribution. Therefore, the uncertainty is at least implicitly available.
However, it should be noted that the number of parameters describing the model here, namely the network weights, is potentially \emph{much} larger than in the approach taken by \citet{domhan2015speedingup}.

\citet{wistuba2019inductive} propose to use biased matrix factorisation to model $\learningcurve(\cdot, \cdot)$.
The approach is settled in the context of neural architecture search.
Knowledge from previous datasets and different architectures is used to estimate the performance of new architectures on the target dataset, and this estimate is used to drive a Bayesian optimisation approach.

\begin{table}[ht!]
    \centering
    \caption{Overview of the discussed learning curve approaches, ordered along the most general question they address, the learning curve type and the data resources used (4 columns).
    In the header, `LC' stands for learning curve, `DS' stands for dataset, and `AL' stands for algorithm. 
    All of them use partial empirical curves on the target dataset of the current learner(s).
    Estimate type: p,r,d are \emph{point}, \emph{range} and \emph{distribution} estimates, respectively. 
    \label{tab:overview}}
    \bgroup
    \setlength{\tabcolsep}{2pt}
   
\begin{tabularx}{\textwidth}{llllllllX}
\toprule
                                           Question &  Type & \rot{LC other DS} & \rot{DS Meta-Feat.} & \rot{LC other AL} & \rot{AL Meta-Feat.} & \rot{Utility} & \rot{Estimate Type} &                                                                                                                                                                                                                             Contributions \\
\midrule
$\refperformance > \tau$ & obs. & \xmark & \xmark & \xmark & \xmark & \xmark & p & \cite{petrak2000fast,vandenbosch2004wrappedprogressivesampling,jamieson2016nonstochasticbestarm,zeng2017progressivesampling} \\
$\refperformance > \tau$ & both & \xmark & \xmark & \xmark & \xmark & \xmark & p & \cite{li2017hyperband} \\
$\pi_{\learner \in \learnerspace}$ & obs. & \cmark & \xmark & \cmark & \xmark & \xmark & p & \cite{leite2005predictingrelativeperformance,leite2007aniterativeprocess,vanrijn2015fastalgorithmselection} \\
$\pi_{\learner \in \learnerspace}$ & obs. & \cmark & \cmark & \xmark & \xmark & \xmark & p & \cite{leite2008selecting,leite2010activetesting} \\
$\pi_{\learner \in \learnerspace}$ & both & \cmark & \cmark & \xmark & \xmark & \xmark & p & \cite{ruhkopf2022masif} \\
$\satpoint$ & iter. & \xmark & \xmark & \xmark & \xmark & \xmark & p & \cite{bishop1995regularization} \\
$\satpoint$ & obs. & \xmark & \xmark & \xmark & \xmark & \xmark & p & \cite{john1996staticvsdynamic,provost1999efficientprogressivesampling,ng2006progressiveoversampling} \\
$\satpoint$ & obs. & \cmark & \xmark & \xmark & \xmark & \xmark & p & \cite{leite2003improvingprogressivesampling,leite2004improvingprogressivesampling} \\
$\limperformance$ & obs. & \xmark & \xmark & \xmark & \xmark & \xmark & p & \cite{cortes1994limits} \\
$\economicstoppingpoint$ & obs. & \xmark & \xmark & \xmark & \xmark & \xmark & p & \cite{meek2002thelcsamplingmethod} \\
$\economicstoppingpoint$ & obs. & \xmark & \xmark & \xmark & \xmark & \cmark & p & \cite{weiss2006maximizingclassifierutility,weiss08maximizingclassifierutility} \\
$\learningcurve(\learner, |\trainset|)$ & obs. & \cmark & \xmark & \xmark & \xmark & \xmark & r & \cite{chandrashekaran2017speedinguphpo} \\
$\learningcurve(\learner, |\trainset|)$ & iter. & \xmark & \xmark & \cmark & \cmark & \xmark & p & \cite{baker2018acceleratingnas} \\
$\underline{\learningcurve(\learner, \cdot)}$ & obs. & \xmark & \xmark & \xmark & \xmark & \xmark & p & \cite{sabharwal2016selecting} \\
$\underline{\learningcurve(\learner, \cdot)}$ & obs. & \xmark & \xmark & \xmark & \xmark & \xmark & r & \cite{mohr2023fast} \\
$\learningcurve(\learner, \cdot)$ & iter. & \xmark & \xmark & \xmark & \xmark & \xmark & p & \cite{cortes1993learningcurves} \\
$\learningcurve(\learner, \cdot)$ & obs. & \xmark & \xmark & \xmark & \xmark & \xmark & p & \cite{john1996staticvsdynamic,frey1999powerlawfordecisiontrees,gu2001modelingclassificationperformance,boonyanunta2004predicting,singh2005modeling,hess2010learning,kolachina2012predictioninmachinetranslation,richter2019approximating} \\
$\learningcurve(\learner, \cdot)$ & obs. & \xmark & \xmark & \xmark & \xmark & \xmark & r & \cite{mukherjee2003estimatingdssizerequirements,figueroa2012predictingsamplesize,koshute2021recommending} \\
$\learningcurve(\learner, \cdot)$ & iter. & \xmark & \xmark & \xmark & \xmark & \xmark & d & \cite{domhan2015speedingup} \\
$\learningcurve(\learner, \cdot)$ & obs. & \cmark & \xmark & \xmark & \xmark & \xmark & p & \cite{cardonaescobar2017efficienthpo} \\
$\learningcurve(\learner, \cdot)$ & both & \cmark & \xmark & \xmark & \xmark & \xmark & d & \cite{adriaensen2023efficient} \\
$\learningcurve(\learner, \cdot)$ & both & \xmark & \xmark & \xmark & \xmark & \xmark & d & \cite{egele2024unreasonable} \\
$\learningcurve(\learner, \cdot)$ & obs. & \cmark & \cmark & \xmark & \xmark & \xmark & p & \cite{kielhofer2024learning} \\
$\utilitycurve(\learner,\cdot)$ & obs. & \xmark & \xmark & \xmark & \xmark & \cmark & p & \cite{last2007predicting,last2009improving} \\
$\learningcurve(\cdot, \cdot)$ & both & \xmark & \xmark & \cmark & \cmark & \xmark & d & \cite{swersky2014freezethawbo,klein2017lcpredictionwithbnn} \\
$\learningcurve(\cdot, \cdot)$ & obs. & \xmark & \xmark & \cmark & \cmark & \xmark & d & \cite{klein2017fabolas} \\
$\learningcurve(\cdot, \cdot)$ & both & \cmark & \xmark & \cmark & \cmark & \xmark & d & \cite{wistuba2019inductive} \\
\bottomrule
\end{tabularx}
    \egroup
\end{table}

\section{Summary and Open Research Directions}
\label{sec:conclusions}
Learning curves have been a vital resource for decision making in machine learning for several decades, and they have gained significant attention over the last years.
Learning curves have proven to be a suitable solution for different types of decision-making situations, i.e., data acquisition, early stopping, and early discarding for model selection. 

We have provided a formal definition of various types of learning curves (Sec.~\ref{sec:background}). 
There are two predominant types of learning curves in the machine learning literature, i.e.: the \samplecurve (i.e., the type of learning curve that one obtains when giving a learner more training instances) and the \itcurve (i.e., the type of learning curve that one obtains when allowing an algorithm to process the data multiple times, see for example the number of epochs of a neural network). 
While both types of learning curves seem similar, they have distinct semantic meanings and characteristics.
Both types of learning curves can be extended to a utility curve, which considers the cost of computational resources or data acquisition. 
Additionally, we have contrasted these against other types of (learning) curves, such as feature curves, capacity curves, and curves obtained by data-centric models, such as active learning or curriculum learning. 

We have described the basic concepts of modelling a learning curve (Sec.~\ref{sec:curvemodels}). 
There are various parametric models that incorporate domain knowledge about what we already know about the shape of learning curves (e.g., the three-parameter inverse power law-model). 
Even when the performance of a learner is observed at very few anchors, the learning curve can already be extrapolated to make predictions about larger anchors. 
Additionally, one can decide to also model a degree of uncertainty, either as a range estimate or as a distribution. 
We distinguish between two types of uncertainty, i.e., epistemic uncertainty and aleatoric uncertainty, and relate these concepts to the literature on modelling learning curves. 
Typically, when uncertainty is being modelled, the epistemic uncertainty is being modelled, but in some cases, the aleatoric uncertainty is being modelled~\citep[see, e.g.,][]{klein2017lcpredictionwithbnn}.

We have provided a unified framework for methods that utilise learning curves for decision making in machine learning (Sec.~\ref{sec:taxonomy}). 
This framework categorises these methods along three axes: the decision situation that they address, the questions that can be addressed with learning curves, and the data resources that can be used to model the learning curves. 
Notably, Fig.~\ref{fig:questions:learningcurve} shows an overview of all questions that can be addressed by learning curves. 
There are various ways to address decision situations with learning curves;
for example, questions about the saturation point of a given learner or whether a learner will perform better than another learner at a given amount of data. 
These questions can be further generalised, eventually ranging in complexity from binary questions to questions that address how any learner behaves at any budget. 

We have done an extensive literature survey, categorising all learning curve methods that we are aware of into this framework (Sec.~\ref{sec:literature}). 
Table~\ref{tab:overview} shows an overview of the methods we have discussed in this survey, contextualising them according to these criteria. 
This table can be seen as an extension of Fig.~\ref{fig:problem_solution}.
Based on this literature survey, we describe several directions for future work. 

{\bf More experimental databases for learning curve research to support the full complexity of learning curve methods.}
Doing relevant research on learning curves requires extensive computing power. 
Exploring a \samplecurve inherently requires many models to restart the learning process at different anchors, whereas exploring an \itcurve is often done on neural networks that come with their distinct layers of complexity~\citep[see, e.g.,][]{white2023neural}.
A common way to address this is by experimental databases or surrogate benchmarks that store certain experimental results. 
This allows for fast experimentation and, therefore, faster development cycles. 
While several of these experimental databases for learning curves exist (as outlined in Sec.~\ref{sec:background:empiricalcurves}), these currently do not capture the full scale of learning curves resources or questions that can be answered using learning curves. 

{\bf A quantitative benchmark on learning curve extrapolation methods.}
Many models have been proposed to extrapolate learning curves and make predictions about the performance of a learning at a higher budget~\citep[see, e.g.,][]{gu2001modelingclassificationperformance}. 
However, these models have only been subject to limited comparison. 
While \citet{gu2001modelingclassificationperformance} compared various parametric models against each other, and \citet{kielhofer2024learning} compared a representative parametric model against a representative metalearning model across many different settings, more research is needed. 
Fig.~\ref{fig:citations:withoutlcmodel} and Fig.~\ref{fig:citations:withlcmodel} already show that, while many papers are aware of other methods and cite those methods (grey arrows), only very few actively compare against each other (blue arrows). 
Moreover, many of these learning curve models are used as a small component in a larger system, e.g., an AutoML system.
In such a case, the predictive performance of the learning curve model might not even be measured, as eventually, one often measures the quality of the complete system; in the case of an AutoML system, the performance of the final selected model. 
Due to this modular nature, improvements on the learning curve extrapolation method would then be orthogonal to improvements on the AutoML system. 

{\bf Tighter integration of learning curve extrapolation methods with AutoML systems.}
We already noted a clear opportunity for AutoML systems in Sec.~\ref{sec:decisionsituations:earlydiscarding}.
In situations where multiple learners are being compared against each other, the training set needs to be further split into an actual training set and a validation set to select the best learner to be tested on the test set (which can only be seen once). 
The existence of this validation set already shows an opportunity for learning curve methods; while a learner is being selected based on its performance after being trained on the split-off training set, what is relevant is its performance after being trained on the original training set (i.e., the split-off training set plus validation set). 
It is not necessarily the case that the same learner performs best on both. 
Learning curve extrapolation models can predict which learner will eventually perform best on an anchor of the size of the original training set. 

{\bf More learning curve methods that operate on low-level questions.}
A prominent question that arises from the literature survey is whether simple questions can be treated more simplistically.
Fig.~\ref{fig:questions:learningcurve} shows four binary questions. While approaches aim to answer these questions, most do so by implicitly answering a more difficult question (see Table~\ref{tab:overview} and Fig.~\ref{fig:problem_solution}).
Some of these questions are really at the core of the discussed approaches.
For example, will the learning curve intersect with the learning curve of the currently best model?
Most approaches create a learning curve model for this, implicitly solving a much more difficult problem.
While those models, if appropriate, have the potential to provide additional interesting insights, the question arises whether simpler approaches could reliably solve those problems while needing much less online data.
Approaches that remain faithful to this question level (such as successive halving and hyperband) have proven effective and received considerable attention.
Additionally, we see a clear opportunity for incorporating uncertainty into methods that address the binary question, effectively providing the chance that a particular learner will be better than another learning at a given budget. 

{\bf A learning curve method that makes use of all types of data resources.}
Learning curve methods can make use of various resources to model the learning curve (see Fig~\ref{fig:resources}). 
For example, the inverse power law model uses the current learner's learning curve on the same dataset. In contrast, metalearning models often also make use of learning curves of either the current learner or other learners on other datasets \citep[see, e.g.,][]{leite2005predictingrelativeperformance}. 
A reasonable assumption is that the methods that utilise more types of data resources would be more accurate. 
Table~\ref{tab:overview} reveals that no learning curve model utilises all types of data resources. 
Indeed, combining anchors across learners and datasets might be a complex task, but when this is done successfully, it will enormously increase our understanding of learning curves.

\subsection*{Acknowledgements.} We thank Pavel Brazdil, Isabelle Guyon, Aaron Klein, and Marco Loog for their insightful discussions and feedback on this survey.

\clearpage
\appendix

\section{Notation}
\label{app:notation}
Table~\ref{tab:notation} contains an overview of the notation used throughout this paper. 


\begin{table}[ht!]
\caption{Overview of notation. \label{tab:notation}}
\begin{tabularx}{\textwidth}{lX}
   \hline
   term & description \\
   \hline
  \dataspace & The space of all possible datasets \\
  \dataset & An instantiation of a dataset \\
  \trainset & The instances from a given dataset based upon which a given hypothesis $h$ is trained \\
  \instancespace & The space of all possible input values for a given dataset $d$\\
  \labelspace & The space of all possible labels of a given dataset $d$\\
  \hypospace & The space that a given model or hypothesis $h$ can take \\
  $h$ & Model or hypothesis $h$, induced based on a given train set \trainset \\
  \learner & An algorithm that given a training set \trainset induces a hypothesis $h$ \\
  \learnerspace & Set of all possible learners under consideration \\
  $\risk_{out}$ & The theoretical performance of a hypothesis under the true distribution of the data (note that this true distribution is typically unknown) \\
  $\risk_{in}$ & The empirical risk of a hypothesis under some sample \dataset (i.e., a dataset) from the true distribution \\
  \learningcurve(\learner, \sampleanchor) & true mean performance of learner \learner when trained on \sampleanchor samples (related to observation learning curve) \\
  \learningcurve(\learner,\sampleanchor,\timeanchor) & true mean performance of learner \learner when trained on \sampleanchor samples with \timeanchor iterations (related to iteration learning curve) \\
  \lcab & The performance of a given learner \learner trained on a dataset of sample size \budget (in case of observation curve) or iteration \budget (in case of iteration curve). In contrast to \learningcurve. \lcab is a \emph{random variable} with \learningcurve as mean value\ \\
  \curvemeanab & The mean of \lcab \\
  \curvenoiseab & The variance of \lcab \\
  \lcobservationset & Set of performance observations from anchors of historic learning curves, possibly for different learners, e.g., $\lcobservationset = \{(\learner_1, \budget_1), (\learner_2,\budget_2), (\learner_3,\budget_3), \ldots \}$, where $\learner_i$ are learners and $b_j$ are budgets. \\
  \lcestab & The performance estimated by a learning curve model for learner \learner at budget \budget. Maybe a point, range, or distributional estimate.\\
  \lcestb & The performance estimated by a learning curve model for learner \learner at budget \budget if the model does not generalize across learners. Maybe a point, range, or distributional estimate.\\
  \sampleanchor & anchor size, indicating the size of a subsample of the dataset\\
  \timeanchor & number of iterations, e.g., in the case of neural networks, the number of epochs \\
  \budget & generic symbol for an anchor, stands either for \sampleanchor or \timeanchor, depending on the context. \\
  \satpoint & The anchor size at which the performance of a learner saturates \\
  
  \satperformance & The performance of the learner at the saturation point \\
  \economicstoppingpoint & The anchor size at which the utility (a performance measure divided by a cost measure) is maximized \\
  \refpoint & Used in the framework for the binary question as a threshold on the budget \\
  \refperformance & Used in the framework for the binary question as a threshold on the performance \\
  \params & The parameters of a parametric function, for example, the parameters of the IPL-model are $(\alpha, \beta, \gamma)$\\
\hline
\end{tabularx}
\end{table}

\subsection*{Declarations}
\textbf{Funding.}
Felix Mohr was supported through the project  ING-312-2023 from Universidad de La Sabana.

\smallskip
\noindent
\textbf{Conflicts of interest.}
Not Applicable

\smallskip
\noindent
\textbf{Ethics approval.}
Not Applicable

\smallskip
\noindent
\textbf{Consent to participate.}
Not Applicable

\smallskip
\noindent
\textbf{Consent for publication.}
Not Applicable

\smallskip
\noindent
\textbf{Availability of data and material.}
Not Applicable

\smallskip
\noindent
\textbf{Code availability.}
Not Applicable

\smallskip
\noindent
\textbf{Authors' contributions.}
Both authors were involved in the redaction of all parts of the document. F.M. had the lead in all sections; J.N.v.R. contributed to all sections by actively writing and critically reviewing the text. 

\bibliography{sn-bibliography}

\end{document}